\documentclass[10pt,journal,compsoc]{IEEEtran}
\usepackage{hyperref}       
\usepackage{url}            
\usepackage{booktabs}       
\usepackage{amsfonts}       
\usepackage{nicefrac}       
\usepackage{microtype}      
\usepackage{color}

\usepackage{amsmath}
\usepackage{amssymb}

\usepackage{graphicx} 
\usepackage{algorithm}
\usepackage{algorithmic}
\usepackage{subfigure}

\usepackage{diagbox}
\usepackage{multirow}
\usepackage{bigstrut}

\usepackage{amsmath,amsfonts,amsthm,amssymb}

\newcommand{\mU}{\mathcal{U}}

\newcommand{\mR}{\mathcal{R}}
\newcommand{\mD}{\mathcal{D}}
\newcommand{\mP}{\mathcal{P}}
\newcommand{\mC}{\mathcal{C}}

\newcommand{\mH}{\mathbb{H}}

\newcommand{\ep}{\mathbb{E}}

\usepackage{amssymb}
\usepackage{pifont}
\newcommand{\cmark}{\ding{51}}%
\newcommand{\xmark}{\ding{55}}%

\newtheorem{theorem}{Theorem}[section]
\newtheorem{corollary}{Corollary}[theorem]
\newtheorem{lemma}[theorem]{Lemma}

\newtheorem*{theorem3}{Theorem 3.3}
\newtheorem*{lemma1}{Lemma 3.1}
\newtheorem*{lemma2}{Lemma 3.2}
\newtheorem*{corollary2}{Corollary 3.2.1}
\newtheorem*{corollary3}{Corollary 3.3.1}

\ifCLASSOPTIONcompsoc
  \usepackage[nocompress]{cite}
\else
  \usepackage{cite}
\fi
\ifCLASSINFOpdf
\else
\fi
\begin{document}
%
\title{Triple Generative Adversarial Networks}
%
%
%
%

\author{Chongxuan~Li,
        Kun~Xu,
        Jiashuo~Liu,
        Jun~Zhu,~\IEEEmembership{Member, IEEE,}
        and~Bo~Zhang
\IEEEcompsocitemizethanks{
\IEEEcompsocthanksitem C. Li, K. Xu, J. Liu, J. Zhu and B. Zhang are with Department of Computer Science
and Technology; TNList Lab; Institute for AI; Center for Bio-Inspired Computing
Research, Tsinghua University, Beijing, 100084 China. $^\dag$J. Zhu is the corresponding author.
Email: chongxuanli1991@gmail.com;
kunxu.thu@gmail.com; liu-js16@mails.tsinghua.edu.cn;
dcszj@tsinghua.edu.cn;
dcszb@tsinghua.edu.cn.}}

%
%

\markboth{Journal of \LaTeX\ Class Files,~Vol.~14, No.~8, August~2015}%
{Shell \MakeLowercase{\textit{et al.}}: Bare Demo of IEEEtran.cls for Computer Society Journals}
%



\IEEEtitleabstractindextext{%
\begin{abstract}
We propose a unified game-theoretical framework to perform classification and conditional image generation given limited supervision. It is formulated as a three-player minimax game consisting of a generator, a classifier and a discriminator, and therefore is referred to as Triple Generative Adversarial Network (Triple-GAN). The generator and the classifier characterize the conditional distributions between images and labels to perform conditional generation and classification, respectively. The discriminator solely focuses on identifying fake image-label pairs. Under a nonparametric assumption, we prove the unique equilibrium of the game is that the distributions characterized by the generator and the classifier converge to the data distribution. As a byproduct of the three-player mechanism, Triple-GAN is flexible to incorporate different semi-supervised classifiers and GAN architectures. We evaluate Triple-GAN in two challenging settings, namely, semi-supervised learning and the extreme low data regime. In both settings, Triple-GAN can achieve excellent classification results and generate meaningful samples in a specific class simultaneously. In particular, using a commonly adopted 13-layer CNN classifier, Triple-GAN outperforms extensive semi-supervised learning methods substantially on more than 10 benchmarks no matter data augmentation is applied or not.
\end{abstract}

}

\maketitle

\IEEEdisplaynontitleabstractindextext

%
\IEEEpeerreviewmaketitle

\IEEEraisesectionheading{\section{Introduction}\label{sec:introduction}}

%
%
%
%

\IEEEPARstart{G}{enerative} adversarial networks (GANs)~\cite{goodfellow2014generative} have made significant progress in generating realistic images~\cite{radford2015unsupervised,arjovsky2017wasserstein,karras2017progressive,karras2018style,brock2018large} and learning representations~\cite{donahue2016adversarial,dumoulin2016adversarially,donahue2019large} in an unsupervised manner, which makes GANs popular in many applications~\cite{isola2017image,zhu2017unpaired,zhang2017stackgan,ledig2017photo, donahue2019large}. 
The paradigm of GAN is a two-player game, where a generator $G$ takes a random noise $z$ as input and produces a fake data sample and a binary discriminator $D$ 
identifies whether a certain sample is true or fake. The learning objective of $G$ is to fool $D$. Formally, it can be formulated as follows:
\begin{eqnarray}\label{eq:gan}
\min_{G}\max_{D} \ep_{p(x)} [\log(D(x))] + \ep_{p_z(z)}[\log (1 - D(G(z)))],
\end{eqnarray}
where $p(x)$ and $p_z(z)$ are the data distribution and the prior distribution of the random noise $z$, respectively.
Theoretically, such a two-player game is sufficient for unsupervised learning. Indeed, the distribution defined by $G$ will converge to the data distribution in the equilibrium of the game, under a nonparametric assumption~\cite{goodfellow2014generative}.

Though promising and appealing, the representations learned by purely unsupervised GANs are not sufficiently good in some down-stream tasks such as classification and conditional generation with disentangled factors, which are of general interests~\cite{krizhevsky2012imagenet,simonyan2014very, szegedy2015going,he2016deep,yang2015weakly,chen2016infogan,li2018graphical}.
In fact, on one hand, the representations learned by unsupervised deep generative models (DGMs) (including GANs) can be less discriminative than those learned by supervised deep neural networks (DNNs) and the predictions made based on the representations are not sufficiently accurate~\cite{li2015max}. On the other hand, unsupervised learning disentangled factors of clear physical meaning and generating samples according to the given factors are challenging and highly depend on the inductive biases of the model~\cite{locatello2018challenging}. 

A natural way to address these issues is to incorporate label information into GANs~\cite{mirza2014conditional,springenberg2015unsupervised,odena2016semi,salimans2016improved,dai2017good,zhang2018dada}. Because supervision can be rare in many scenarios~\cite{kingma2014semi,lake2015human}, we investigate the ones where only a small number of labels are accessible in this paper. In particular, we consider two closely related yet challenging settings.
The first one is {\it semi-supervised learning}~\cite{springenberg2015unsupervised,odena2016semi,salimans2016improved}, where a large amount of unlabeled data is available and can be explored to boost the classification performance (as well as the conditional generation performance in DGMs~\cite{kingma2014semi}). 
The second one is the more challenging {\it extreme low data regime}~\cite{zhang2018dada}, where no unlabeled data is available. 

\begin{figure}[t]
\centering
\includegraphics[width=0.98\columnwidth]{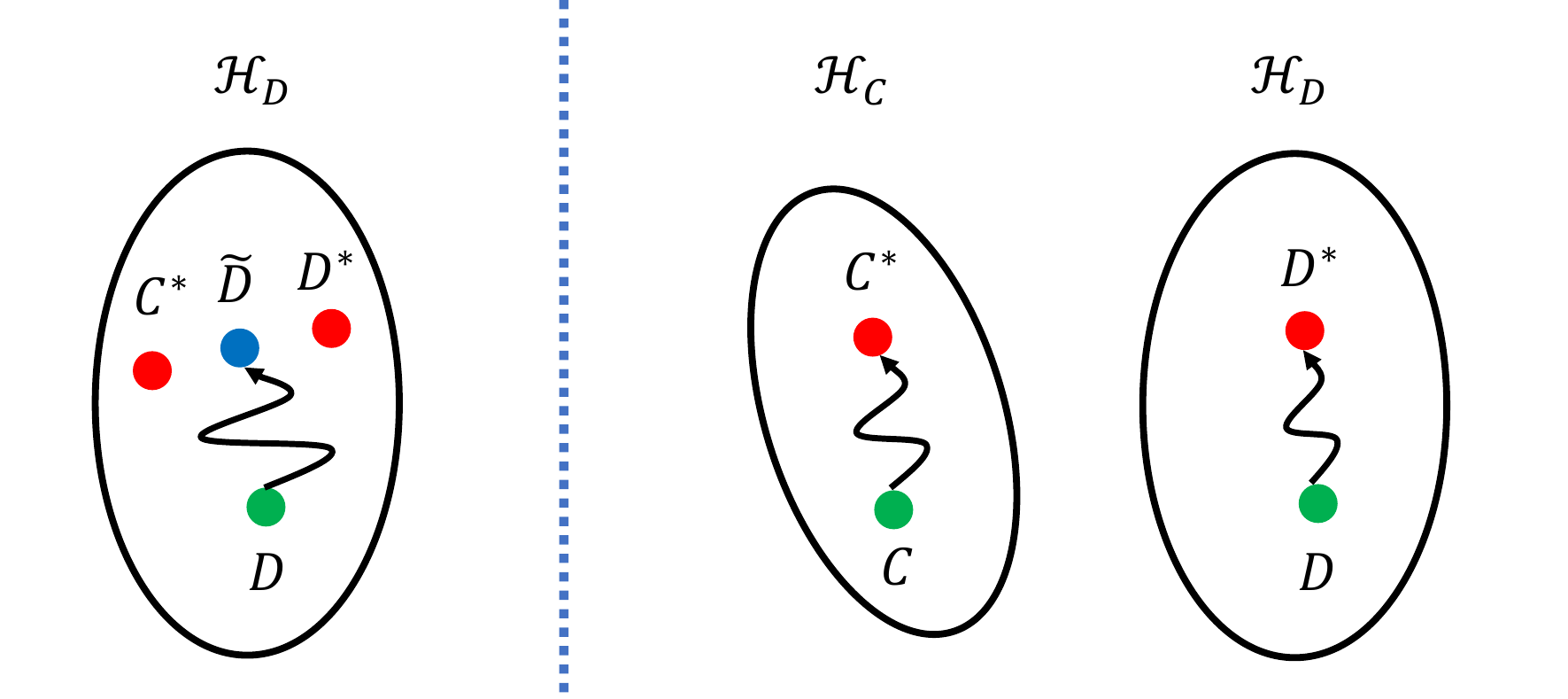}
\vspace{-.1cm}
\caption{The left panel illustrates a representative prior work~\cite{springenberg2015unsupervised}. $D$ is the initialized model in the hypothesis space $\mathcal{H}_D$. $C^*$ is the optimal classifier and $D^*$ is the optimal discriminator when $p_g(x)=p(x)$. $D$ will converge to a solution 
$\tilde{D}$ that depends on the trade-off between classification and generation. The right panel illustrates our approach where $C$ and $D$ are optimized in two separated hypothesis spaces $\mathcal{H}_C$ and $\mathcal{H}_D$, respectively. $C$ and $D$ will converge to the corresponding optimum under a nonparametric assumption~\cite{goodfellow2014generative} (See Sec.~\ref{sec:theory} for the proof). 
}
\vspace{-.2cm}
\label{fig:triple-motivation}
\end{figure}

Several methods~\cite{springenberg2015unsupervised,odena2016semi,salimans2016improved,zhang2018dada} are proposed to solve the classification task by introducing a categorical discriminator, which also serves as a classifier, under the two-player game formulation in Eqn.~\eqref{eq:gan}. The discriminator is trained to identify fake samples and predict labels simultaneously.
Though obtaining a substantial improvement over DNNs in classification, these methods lack a theoretical analysis of the equilibrium of the game.
We take CatGAN~\cite{springenberg2015unsupervised} as a representative example (See Fig.~\ref{fig:triple-motivation} for an illustration). The discriminator in CatGAN maximizes the predictive entropy of a fake sample and minimizes that of a real sample. 
Assume that $p_g(x)$, the distribution defined by $G$, converges to the data distribution $p(x)$ and we have $x\sim p_g(x) = p(x)$. On one hand, when $G$ converges, the optimal discriminator $D^*$ cannot distinguish whether $x$ is real or fake~\cite{goodfellow2014generative} and therefore has high predictive entropy. On the other hand, the optimal classifier $C^*$ should classify $x$ correctly and has low predictive entropy. Thus, generally, $C^*$ and $D^*$ are different points in the same hypothesis space and the discriminator will converge to a solution depending on the trade-off between classification and generation. The empirical results in Improved-GAN~\cite{salimans2016improved} well support our analysis. In fact, Improved-GAN proposes two alternative training objective functions. The first  {\it feature matching} objective works well in classification but fails to generate indistinguishable samples. 
The second {\it minibatch discrimination} objective is good at image generation with full labels but cannot perform classification accurately~\cite{salimans2016improved}.
Therefore, we argue that the two-player formulation may lead to a sub-optimal classifier.

As for the class-conditional generation task, conditional GAN~\cite{mirza2014conditional} and its variants~\cite{radford2015unsupervised,dumoulin2016adversarially,odena2016conditional,zhang2018dada}  can train a conditional generator given fully labeled data (See details in Sec.~\ref{sec:background}). 
However, how to leverage unlabeled data in the conditional generation task is highly nontrivial due to the presence of missing labels in semi-supervised learning. 
In fact, to our best knowledge, none of the existing GANs~\cite{springenberg2015unsupervised,odena2016semi,salimans2016improved} can exploit partially labeled data to generate samples in a specific class. 
Again, we believe that the problem is intrinsically caused by the two-player formulation.
Specifically, the discriminators in these methods estimate a single data instead of a data-label pair and the label information is totally ignored.
Therefore, the generators will not receive any learning signal regarding the label information from the discriminators and hence cannot control the semantics of the generated samples, which is not satisfactory.

In this paper, we propose the {\it triple generative adversarial network} (Triple-GAN) framework for both classification and class-conditional image generation with limited supervision. Specifically, we introduce two conditional networks---a classifier and a generator to generate fake labels given real data and fake data given real labels, which will perform the classification and class-conditional generation tasks respectively.
To jointly justify the quality of the samples from the conditional networks, we define a discriminator network which has the sole role of distinguishing whether a data-label pair is from the real labeled dataset or not. The resulting model is called Triple-GAN because we consider three networks as well as three joint distributions, i.e. the true data-label distribution and the distributions defined by the two conditional networks.
Directly motivated by the desirable equilibrium, we carefully design compatible objective functions, including adversarial losses and cross-entropy losses, for the three players. The objective functions are optimized by alternative stochastic gradient descent~\cite{goodfellow2014generative}.

Theoretically, we prove that the unique equilibrium of Triple-GAN 
is that the distributions defined by both the generator and the classifier converge to the data distribution under a nonparametric assumption~\cite{goodfellow2014generative}. This equilibrium implies that the classifier of Triple-GAN can potentially converge to a better solution than existing work~\cite{mirza2014conditional,radford2015unsupervised,dumoulin2016adversarially,odena2016conditional}.  Though our implementation does not rigorously follow the theoretical analysis, it does provide valuable insights. Besides, the discriminator can access the label information of the unlabeled data from the classifier and then force the generator to generate correct image-label pairs, which solves the class-conditional generation task.

Empirically, we notice that recent advances in semi-supervised classification~\cite{tarvainen2017mean} and GANs~\cite{miyato2018spectral,miyato2018cgans}
are built upon very different network architectures and loss functions.
As a byproduct of decoupling the hypothesis spaces of the classifier and discriminator, it is convenient to adopt these advances simultaneously in Triple-GAN without considering the trade-off between classification and generation.
In fact, we implement two instances, called as {\it Triple-GAN-V1} and  {\it Triple-GAN-V2} respectively.
In Triple-GAN-V1, for a fair comparison, we use models that are comparable to the GAN-based methods~\cite{springenberg2015unsupervised,odena2016semi,salimans2016improved,zhang2018dada} to validate our motivation and theoretical results. In Triple-GAN-V2, we employ a mean teacher classifier~\cite{tarvainen2017mean} and a projection discriminator~\cite{miyato2018cgans} with spectral normalization~\cite{miyato2018spectral}, to compete against a large family of strong baselines.
Both instances can achieve excellent classification results and generate meaningful samples in a specific class simultaneously, evaluated on the widely adopted SVHN~\cite{netzer2011reading}, CIFAR10~\cite{krizhevsky2009learning} and Tiny ImageNet\footnote{Available at https://tiny-imagenet.herokuapp.com/} datasets in semi-supervised learning and the extreme low data regime. 
In particular, using a commonly adopted 13-layer CNN classifier, Triple-GAN outperforms extensive semi-supervised learning methods~\cite{laine2016temporal,gan2017triangle,dai2017good,luo2018smooth,oliver2018realistic,lee2013pseudo,miyato2015distributional,tarvainen2017mean, iscen2019label,athiwaratkun2018there} substantially on more than 10 benchmarks no matter data augmentation is applied or not. Besides, with only $8\%$ labels, it achieves comparable Inception Score (IS)~\cite{salimans2016improved} and Fr\'echet Inception Distance (FID)~\cite{heusel2017gans}  to the strong baseline CGAN-PD~\cite{miyato2018cgans} trained with full labels on the CIFAR10 dataset.

Overall, our main contributions are:
\begin{enumerate}
    \item We propose a unified GAN-based framework consisting of three players (thus named Triple-GAN) to perform classification and conditional generation given limited supervision;
    \item We prove that, with carefully designed objective functions, Triple-GAN leads to a unique desirable equilibrium under a nonparametric assumption~\cite{goodfellow2014generative};
    \item We show that Triple-GAN can significantly outperform a large number of recent semi-supervised learning methods on more than 10 benchmarks using a commonly adopted 13-layer CNN classifier;
    \item We demonstrate that, given partially labeled data,  Triple-GAN is able to disentangle category from style features and get samples of comparable quality to the strong baseline~\cite{miyato2018cgans} with full labels.
\end{enumerate}


\section{Background}
\label{sec:background}

In this section, we present several discriminative semi-supervised learning methods adopted in Triple-GAN and existing GAN variants with supervision. We will
discuss other related work comprehensively in Sec.~\ref{sec:related_work}.

\subsection{Discriminative semi-supervised learning methods}
\label{sec:ssl_bg}

There are extensive discriminative approaches in semi-supervised learning. Such methods define different regularization terms on the unlabeled data. 

\textbf{Entropy regularization.} Entropy regularization~\cite{yves2004semi} minimizes the predictive entropy of the unlabeled data, which is defined as follows:
\setlength{\arraycolsep}{0.0em}
\begin{align}\label{eq:entropy}
\mathcal{R_U} = - \ep_{p(x)}\ep_{p_c(y|x)}[\log p_c(y|x)].
\end{align}

\textbf{Consistency regularization.} Consistency regularization~\cite{laine2016temporal} minimizes the mean square error (MSE) between two predictions of the same unlabeled data given different noises like dropout masks. Formally, it is defined as follows:
\begin{equation}
\label{eq:consist}
    \mR_{\mU} = \ep_{p(x)}||p_c(y|x, \epsilon) - p_c(y|x, \epsilon')||^2,
\end{equation}
where $||\cdot||^2$ is the square of the $l_2$-norm and $\epsilon$ and $\epsilon'$ denote two different random noises. 
We note that there are other optional loss functions such as the KL-divergence in Eqn.~\eqref{eq:consist}. However, according to the empirical study in prior work~\cite{tarvainen2017mean}, MSE performs best. Though the exact reason is not clear, intuitively, MSE may avoid over-confident predictions in the consistency loss compared to the KL divergence.

\textbf{Mean teacher.} Mean teacher~\cite{tarvainen2017mean} maintains an exponential moving average of the classifier as a {\it teacher} and enforces the predictions made by the classifier and the teacher on the same data to be the same. Formally, it is formulated as:
\begin{equation}
\label{eq:mean_t}
    \mR_{\mU} = \ep_{p(x)}||p_c(y|x, \epsilon) - p_t(y|x, \epsilon')||^2,
\end{equation}
where $\epsilon$ and $\epsilon'$ denote two different random noises and $p_t(y|x,\epsilon')$ denotes the prediction of the teacher model. The reason of using the MSE instead of the KL divergence is the same as in Eqn.~\eqref{eq:consist}. Mean teacher is a strong baseline and many recent methods~\cite{luo2018smooth,oliver2018realistic,iscen2019label,athiwaratkun2018there} are built upon it.

\subsection{GANs with supervision}

Existing GANs with supervision adopt a similar two-player formulation and lacks a theoretical analysis of the equilibrium.

\textbf{CatGAN.} CatGAN~\cite{springenberg2015unsupervised} extends the original GAN to semi-supervised learning by employing a categorical discriminator, which also serves as a classifier.
The learning problem of CatGAN is formulated as:
\setlength{\arraycolsep}{0.0em}
\begin{align}\label{eq:catgan}
\min_G \max_D \:\:& \lambda \mathbb{E}_{p(x, y)}[\log p_c(y|x)] + \ep_{p(x)}\ep_{p_c(y|x)}[\log p_c(y|x)] \nonumber \\
 & -\mathbb{E}_{p_z(z)}\ep_{p_c(y|G(z))}[\log p_c(y|G(z))],
\end{align}
where $p_c(y|x)$ is the predictive distribution defined by the classifier and $\lambda$ is a hyperparameter.
The first term is the traditional cross-entropy loss on the labeled data for classification.
The second term is the entropy regularization~\cite{yves2004semi} presented in Eqn.~\ref{eq:entropy}.
The last term is the generalized adversarial losses on fake data generated from $G$.

\textbf{Improved-GAN.} Improved-GAN~\cite{salimans2016improved} shares the same settings with CatGAN~\cite{springenberg2015unsupervised} but uses a categorical discriminator that has one extra class that corresponds to the fake samples. The learning problem for the discriminator is: 
\begin{eqnarray}\label{eq:improved_d}
\max_D &&\ep_{p(x, y)} [\log p_c(y|x, y < K + 1)] \nonumber\\
&+&  \ep_{p_z(z)} [\log p_c( y=K+1|G(z))] \nonumber\\
&+& \ep_{p(x)} [\log(1 - p_c(y=K+1|x))],
\end{eqnarray}
where $K$ is the number of real classes and the $(K+1)$-th class corresponds to the fake samples. The generator does not directly fool the discriminator but has two alternative objective functions. In this paper, we mainly compare to {\it feature matching}, which enforces $G$ to produce samples that have similar features (extracted by a neural network) to the real samples, which is formulated as:
\begin{eqnarray}\label{eq:improved_g}
\min_G  ||\ep_{p(x)}[f(x)] - \ep_{p_z(z)}[f(G(z))]||_2^2,
\end{eqnarray}
where $f(x)$ is the activation of certain layer in the discriminator given input $x$ and $||\cdot||_2$ is the $l_2$ norm. 

\textbf{DADA.} DADA~\cite{zhang2018dada} considers a more challenging extreme low data regime setting, where only a small number of labeled data is available. DADA also employs a two-player formulation and extends the discriminator to $2K$ classes, where the first $K$ classes are real and the last $K$ classes are fake. DADA proposes a two-phase training schedule. In the first phase, the learning objective of the discriminator is:
\begin{eqnarray}\label{eq:dada_d}
\max_D &&\ep_{p(x, y)} [\log p_c(y|x, y < K + 1)] \nonumber\\
&+&  \ep_{p_z(z)p(y)} [\log p_c(y | G(z, y), K < y < 2K+1)],
\end{eqnarray}
and the learning objective of the generator is:
\begin{eqnarray}\label{eq:dada_g}
\min_G && - \ep_{p(x, y)} [\log p_c(y - K|x, K <y < 2K + 1)] \nonumber\\
&+&  \lambda ||\ep_{p(x, y)}[f(x|y)] - \ep_{p_z(z)p(y)}[f(G(z, y) | y)]||_2^2 ,
\end{eqnarray}
where the last term is a conditional variant of the feature matching loss in Eqn.~\eqref{eq:improved_g} and $
\lambda$ is a hyperparameter. 
In the second phase, the generator is fixed as a data provider and the classifier is trained for classification based on both the real data and fake data.

The two-player framework in prior methods~\cite{springenberg2015unsupervised,salimans2016improved,zhang2018dada} for semi-supervised learning and extreme low data regime can potentially learn a sub-optimal classifier because of the trade-off between classification and generation (See Fig.~\ref{fig:triple-motivation} for illustration and~\cite{salimans2016improved} for empirical evidence). 

\textbf{CGAN.} The conditional GAN (CGAN)~\cite{mirza2014conditional} and its variants~\cite{radford2015unsupervised,dumoulin2016adversarially,odena2016conditional,miyato2018cgans} aim to learn a conditional generator $p_g(x|y)$ to approximate $p(x|y)$. CGAN is very similar to GAN but incorporates label information in both the generator and the discriminator. The learning problem is defined as:
\setlength{\arraycolsep}{0.0em}
\begin{eqnarray}\label{eq:cgan}
\min_{G} \max_D &\ep&_{p(x, y)} [\log D(x, y)] \nonumber \\
&+&  \ep_{p_z(z)p(y)} [\log (1 - D(G(y, z), y))].
\end{eqnarray}
Recent work~\cite{miyato2018cgans} improves CGAN by using a projection discriminator with spectral normalization~\cite{miyato2018spectral}, and obtains impressive generation results.
However, it can be seen that CGAN-based methods require fully labeled data to train a conditional generator in a ``discriminative'' way and it is non-trivial to leverage unlabeled data by extending CGAN. Besides, existing semi-supervised methods~\cite{springenberg2015unsupervised,odena2016semi,salimans2016improved} lack the ability of conditional generation.

\section{Method}
\label{sec:method}

As discussed above, 
prior GANs with limited supervision~\cite{salimans2016improved,radford2015unsupervised,zhang2018dada} are not problemless: (1) they will  converge to a trade-off solution (between classification and generation) which lacks theoretical analysis; and (2) the learned generators cannot control the semantics of the generated samples. Therefore, we argue that the classification and conditional generation tasks given limited supervision are still largely open.
To this end, we proposed a unified game-theoretical framework {\it triple generative adversarial network} (Triple-GAN), which intrinsically avoid the problems in existing methods.

In particular, we consider two closely related yet challenging learning settings in this paper. The first one is semi-supervised learning~\cite{yves2004semi}, where we have a labeled dataset of small size and an unlabeled one of large size. The second one is the extreme low data regime~\cite{zhang2018dada}, where we only have a labeled dataset of small size. In both settings, we want to predict a label $y$ for an input $x$ (i.e., perform classification) as well as to generate a new sample $x$ conditioned on a label $y$ (i.e., generate samples in a given class).

Triple-GAN is based on the insight that the joint distribution can be factorized in two ways,
namely, $p(x, y) = p(x) p(y | x)$ and $ p(x, y) = p(y) p(x | y)$, and that the conditional distributions $p(y | x)$ and $p(x | y)$ are of interests for classification and class-conditional generation, respectively.
To jointly estimate these conditional distributions, which are characterized by a classifier network and a class-conditional generator network, we define a discriminator network which has the sole role of distinguishing whether a input-label pair is from the true data distribution or the models.
Hence, we naturally extend GAN to Triple-GAN,
a three-player game to characterize the process of classification and class-conditional generation with limited supervision, as detailed below.

\subsection{A game with three players}

Triple-GAN consists of three components: (1) a classifier $C$ that (approximately) characterizes the conditional distribution $p_c(y | x) \approx p(y | x)$; (2) a class-conditional generator $G$ that (approximately) characterizes the conditional distribution in the other direction $p_g(x | y) \approx p(x | y)$; and (3) a discriminator $D$ that distinguishes whether a pair of data $(x, y)$ comes from the true distribution $p(x, y)$. All the components are parameterized as neural networks.
Our desired equilibrium is that the joint distributions defined by the generator and the classifier both converge to the true data distribution. To achieve it, we design a game with compatible objective functions for the three players as follows.

We make a mild assumption that the samples from both $p(x)$ and $p(y)$ can be easily obtained.\footnote{In semi-supervised learning, $p(y)$ is assumed same to the distribution of labels on the labeled data, which is uniform in our experiment.}
In the game, after a sample $x$ is drawn from $p(x)$, $C$ produces a fake label $y$ given $x$ following the conditional distribution $p_c(y|x)$. Hence, the fake input-label pair is a sample from the joint distribution $p_c(x, y) = p(x) p_c(y | x)$. Similarly, a fake input-label pair can be sampled from $G$ by first drawing $y \sim p(y)$ and then drawing $x|y \sim p_g(x|y)$, hence from the joint distribution
$p_g(x, y) = p(y) p_g(x | y)$. For $p_g(x|y)$, we assume that $x$ is transformed by the latent style variables $z$ given the label $y$, namely, $x = G(y, z), z \sim p_z(z)$, where $p_z(z)$ is a simple distribution (e.g., uniform or standard normal).
Then, the fake input-label pairs $(x,y)$ generated by both $C$ and $G$ are sent to the discriminator $D$.
$D$ can also access the input-label pairs from the true data distribution as positive samples. We refer the objective functions in the process as adversarial losses, which can be formulated as:
\setlength{\arraycolsep}{0.0em}
\begin{eqnarray}\label{eq:triple_gan}
\min_{C, G} \max_D &\ep&_{p(x, y)} [\log D(x, y)]
+ \alpha \ep_{p_c(x, y)} [\log (1 - D(x, y))] \nonumber \\
&+& (1-\alpha) \ep_{p_g(x, y)} [\log (1 - D(G(y, z), y))],
\end{eqnarray}
where $\alpha \in (0, 1)$ is a constant that controls the relative importance of classification and generation, and we focus on the balance case by fixing it as $1/2$ throughout the paper. 

The game defined in Eqn.~\eqref{eq:triple_gan} achieves its equilibrium if and only if $p(x, y) = (1-\alpha) p_g(x, y) + \alpha p_c(x, y)$  (See proof in Sec. \ref{sec:theory}). Intuitively, it means that the discriminator balances between the true data distribution and a mixture distribution defined by the generator and the classifier. Further, the equilibrium indicates that
if one of $C$ and $G$ tends to the data distribution, the other will also go towards the data distribution, which avoids the incompatible convergence problem as in previous work~\cite{springenberg2015unsupervised,salimans2016improved,zhang2018dada}. 

However, the adversarial losses defined in Eqn.~\eqref{eq:triple_gan} cannot guarantee that $p(x, y) = p_g(x, y) = p_c(x, y)$ is the unique equilibrium, which is unsatisfactory. A natural solution to the problem is to introduce additional losses to encourage two of the three distributions to be the same.
In fact, we consider to minimize $\mathbb{D}_{KL}(p_c(x, y)|| p(x, y))$ and $\mathbb{D}_{KL}(p_c(x, y)|| p_g(x, y))$\footnote{The KL divergence between $p(x, y)$ and $p_g(x, y)$ is less efficient to compute. Therefore, we omit it here.},
where $\mathbb{D}_{KL}(\cdot || \cdot)$ denotes the KL divergence. 
An advantage of the KL divergence over other divergences is that it has an equivalent form of the cross-entropy loss, which is efficient in computation and effective for classification.
Formally, 
we introduce the cross-entropy loss on the labeled data to $C$ as follows:
\setlength{\arraycolsep}{0.0em}
\begin{eqnarray}\label{eq:ce}
\mR_{\mC} = \ep_{p(x, y)}[-\log p_c(y | x)],
\end{eqnarray}
to minimize $\mathbb{D}_{KL}(p_c(x, y)|| p(x, y))$, and a {\it pseudo discriminative loss} on the generated data to $C$ as follows:
\begin{equation}
\mR_{\mP} = \ep_{p_g(x, y)}[-\log p_c(y | x)].
\label{eq:pesudo}
\end{equation}
Theoretically, we prove that optimizing Eqn.~\eqref{eq:pesudo} with respect to the parameters in $C$ is equivalent to minimizing  $\mathbb{D}_{KL}(p_c(x, y)|| p_g(x, y))$ (See proof in Appendix A). 
Note that $\mathbb{D}_{KL}(p_c(x, y)|| p_g(x, y))$ cannot be optimized directly because its computation involves the unknown likelihood ratio $p_g(x,y)/p_c(x,y)$. 
Intuitively, Eqn.~\eqref{eq:pesudo} treats the samples generated from $G$ as extra labeled data for $C$, which will boost the predictive performance given extremely insufficient supervision. Further, we empirically show that the generated data have a different patterns from the randomly augmented data (e.g., shifted and flipped ones) and provide complementary learning signals  (See Sec.~\ref{sec:experiments} for the empirical evidence). 
Finally, the game with the cross-entropy loss and the pseudo discriminative loss is defined as:
\setlength{\arraycolsep}{0.0em}
\begin{eqnarray}\label{eq:triple_gan_full}
&\!& \min_{C, G}\max_D \ep_{p(x, y)} [\log D(x, y)]
+ \alpha \ep_{p_c(x, y)} [\log (1 - D(x, y))] \nonumber \\
&+& (1\!-\!\alpha) \ep_{p_g(x, y)} [\log (1 - D(G(y, z), y))] \!+\! \mR_{\mC} \!+\! \alpha_{\mP} \mR_{\mP},
\end{eqnarray}
where $\alpha_{\mP} > 0$ is a hyperparameter.
For simplicity, we denote the objective function in Eqn.~\eqref{eq:triple_gan} as  $U(C,G,D)$ and that in Eqn.~\eqref{eq:triple_gan_full} as $\tilde U(C,G,D)$.
It will be proven that the game defined by $\tilde U(C,G,D)$ has a unique equilibrium that both $p_c(x, y)$ and $p_g(x, y)$ converge to $p(x, y)$ in Sec.~\ref{sec:theory}. 

\subsection{Theoretical Analysis}
\label{sec:theory}

\begin{algorithm*}[t]
\begin{algorithmic}
\caption{Minibatch stochastic gradient descent training of Triple-GAN.}\label{algo}
\FOR{number of training iterations}
\STATE \textbullet~Sample a batch of pairs $(x_g, y_g) \sim p_g(x, y)$ of size $m_g$, a batch of pairs $(x_c, y_c) \sim p_c(x, y)$ of size $m_c$ and a batch of the labeled data $(x_d, y_d) \sim p(x, y)$ of size $m_d$.
\STATE \textbullet~Update $D$ by ascending along its stochastic gradient according to Eqn.~\eqref{eq:triple_gan_full}:\\
$$
\nabla_{\theta_d}    \left[ \frac{1}{m_d} (\sum_{(x_d, y_d)}     \log D(x_d, y_d))  +  \frac{\alpha}{m_c}    \sum_{(x_c, y_c)}      \log (1  -  D(x_c, y_c))  +  \frac{1-\alpha}{m_g}     \sum_{(x_g, y_g)}     \log (1  -  D(x_g, y_g)) \right].
$$\\
\STATE \textbullet~Compute the unbiased estimators $\hat \mR_{\mC}$ and $\hat \mR_{\mP}$ in Eqn.~\eqref{eq:ce} of $ \mR_{\mC}$ and $\mR_{\mP}$ in Eqn.~\eqref{eq:pesudo} respectively:
$$
\hat \mR_{\mC} = - \frac{1}{m_d}\sum_{(x_d, y_d) } \log p_c(y_d|x_d),\qquad  \hat \mR_{\mP} = - \frac{1}{m_g}\sum_{(x_g, y_g) }  \log p_c(y_g|x_g).
$$\\
\STATE \textbullet~Compute the unbiased estimator $\hat \mR_{\mU}$ of $\mathcal{R_U}$ according to Eqn.~\eqref{eq:entropy},  Eqn.~\eqref{eq:consist} or  Eqn.~\eqref{eq:mean_t} if required.
\STATE \textbullet~Update $C$ by descending along its stochastic gradient according to Eqn.~\eqref{eq:triple_gan_full}:\\
$$
 \nabla_{\theta_c}\left [ \frac{\alpha}{m_c} \sum_{x_c} \sum_{y \in \mathcal{Y}} p_c(y | x_c) \log (1 - D(x_c, y))
+ \hat \mR_{\mC}  + \alpha_{\mP} \hat \mR_{\mP} + \alpha_{\mU} \hat  \mR_{\mU} \right].$$ \\
\STATE \textbullet~Update $G$ by descending along its stochastic gradient according to Eqn.~\eqref{eq:triple_gan_full}:\\
$$
\nabla_{\theta_g}\left [ \frac{1-\alpha}{m_g} \sum_{(x_g, y_g)} \log (1 - D(x_g, y_g))\right] .
$$\\
\ENDFOR
\end{algorithmic}
\end{algorithm*}

We now provide a formal theoretical analysis of Triple-GAN under a nonparametric assumption~\cite{goodfellow2014generative}. For clarity of the main text, we only present the main results here and defer the proof details to Appendix A. 

First, we would like to show that in the game defined by $U(C, G, D)$, the optimal $D$ given $C$ and $G$ balances between the true data distribution and a mixture distribution of $p_c(x, y)$ and $p_g(x, y)$, as summarized in Lemma~\ref{thm:opt_d}.
\begin{lemma}\label{thm:opt_d}
For any fixed $C$ and $G$, the optimal $D$ of the game defined by the objective function $U(C,G,D)$ is:
\begin{equation}
D_{C, G}^*(x, y) = \frac{p(x, y)}{p(x, y) +
p_{\alpha} (x, y)},
\end{equation}
where $p_{\alpha} (x, y) := (1-\alpha)p_g(x, y) + \alpha p_c(x, y)$ is a mixture distribution for $\alpha \in (0, 1)$.
\end{lemma}

Given $D_{C, G}^*$, we can omit $D$ and reformulate the minimax game with the objective function $U(C,G,D)$ as: $V(C,G) = \max_D U(C, G, D),$ whose minimum is summarized as in Lemma~\ref{thm:opt_dd}.
\begin{lemma}\label{thm:opt_dd}
The global minimum of $V(C, G)$ is achieved if and only if $p(x, y) = p_{\alpha}(x, y) $.
\end{lemma}
The lemma shows that the adversarial losses ensure that if one of $p_c(x, y)$ and $p_g(x, y)$ gets closer to the data distribution after convergence, the other one will also be closer to the data distribution.

We can further show that $C$ and $G$ can at least capture the marginal distributions of data, especially for $p_g(x)$, even there may exist multiple global equilibria, as summarized in Corollary~\ref{thm:mrg}.
\begin{corollary}\label{thm:mrg}
Given $p(x, y) = p_{\alpha} (x, y)$, the marginal distributions of $p(x, y)$, $p_c(x, y)$ and $p_g(x, y)$ are the same, i.e. $p(x) = p_g(x) = p_c(x)$ and $p(y) = p_g(y) = p_c(y)$.
\end{corollary}

Given the above result that $p(x, y) = p_{\alpha}(x, y)$, $C$ and $G$ do not compete as in the two-player based formulation and it is easy to verify that $p(x, y) = p_c(x, y) = p_g (x, y)$ is a global equilibrium point. However, it may not be unique and we should minimize an additional objective to ensure the uniqueness. In fact, this is true for the objective function $\tilde U(C,G,D)$ in problem (\ref{eq:triple_gan_full}), as stated below.
\begin{theorem}
The equilibrium of $ \tilde U(C, G, D)$ is achieved if and only if $p(x, y)=p_g(x, y)=p_c(x, y)$.
\end{theorem}

The conclusion essentially motivates our design of Triple-GAN, as we can ensure that both $p_g(x, y)$ and $p_c(x, y)$ will converge to the true data distribution if the model has been trained to achieve the optimum.

We can further show another nice property of $\tilde U$, which allows us to regularize our model for a more stable and better convergence in practice without changing the global equilibrium, as summarized below.

\begin{corollary}\label{thm:reg}
Adding any divergence (e.g. the KL divergence) between any two of the joint distributions or the conditional distributions or the marginal distributions, to $\tilde U$
as an additional regularization to be minimized, will not change the global equilibrium of $\tilde U$.
\end{corollary}

\subsection{Instances and Optimization}
\label{sec:trick}

Thanks to decoupling the hypothesis spaces of the classifier and discriminator, it is convenient to adopt recent techniques (e.g., architectures and loss functions) of semi-supervised classification and GANs simultaneously in Triple-GAN. In particular, we implement two instances, called {\it Triple-GAN-V1} and {\it Triple-GAN-V2}, for different goals.

\textbf{Triple-GAN-V1.} Our preliminary goal of Triple-GAN-V1 is to validate our motivation and theory by comparing to the generative approaches~\cite{mirza2014conditional,springenberg2015unsupervised,li2017max,odena2016semi,salimans2016improved,dai2017good,zhang2018dada}. Therefore, we use the same classifier and generator architectures as the prior work~\cite{salimans2016improved,zhang2018dada} for a fair comparison. Since properly leveraging the unlabeled data is key to success in semi-supervised learning, it is necessary to regularize $C$ heuristically as in many existing generative methods~\cite{rasmus2015semi,springenberg2015unsupervised,salimans2016improved,li2017max,dai2017good} to make more accurate predictions. Similary, we use the entropy minimization loss in Eqn.~\eqref{eq:entropy} on the MNIST and SVHN datasets and the consistency loss in Eqn.~\eqref{eq:consist} on the CIFAR10 and Tiny ImageNet datasets.
 
\textbf{Triple-GAN-V2.} We further introduce Triple-GAN-V2 to compare with the state-of-the-art methods in discriminative semi-supervised learning~\cite{laine2016temporal,gan2017triangle,dai2017good,luo2018smooth,oliver2018realistic,lee2013pseudo,miyato2015distributional,tarvainen2017mean, iscen2019label,athiwaratkun2018there} and conditional generation~\cite{gulrajani2017improved,miyato2018spectral,miyato2018cgans}. For a fair and direct comparison, on one hand, we use the exact same classifier in the representative mean teacher method~\cite{tarvainen2017mean} (See the loss function in Eqn.~\eqref{eq:mean_t}). On the other hand, as for GANs, we adopt the network architectures used in ~\cite{miyato2018cgans} and a projection discriminator~\cite{miyato2018cgans} with 
spectral normalization~\cite{miyato2018spectral}. 

We mention that adding the extra unlabeled data losses in Sec.~\ref{sec:ssl_bg} to $C$ may introduce some bias to Triple-GAN. However, in practice, we observed that they won't hurt but help the convergence of $G$. This is because the losses only affect $C$ directly and if $C$ can make more accurate predictions (i.e. $p_c(x, y)$ gets closer to $p(x, y)$), then $G$ will become better according to Lemma~\ref{thm:opt_dd}. Though our implementation does not rigorously follow the theoretical analysis, it does provide valuable insights and explain the good results in some sense.

\textbf{Optimization.} Given the objective functions, we optimize the three networks alternatively by stochastic gradient descent (SGD) based methods. 
Let $\theta_c$, $\theta_d$ and $\theta_g$ denote the trainable parameters in $C$, $D$ and $G$, respectively. We need to compute the gradients with respect to all the parameters. The gradients with respect to  $\theta_d$ and $\theta_g$ can be easily obtained. However, the computation of the gradients of $\ep_{p(x)}\ep_{p_c(y|x)} [\log (1 - D(x, y))]$ with respect to $\theta_c$ involves a summation over a discrete random variable $y$, i.e. the class label. Note that directly applying the Monte Carlo method is not feasible because the feedback of the discriminator is not differentiable with respect to $\theta_c$.
Therefore, we directly integrate out the class label and the gradients are as follows:
\begin{equation}
\label{eq:grad_cd}
  \ep_{p(x)} \left [ \nabla_{\theta_c} \sum_{y \in \mathcal{Y}} p_c(y | x) \log (1 - D(x, y))\right],
\end{equation}
where $\mathcal{Y}$ is the set of all possible categories. Note that the calculation in Eqn.~\eqref{eq:grad_cd} only requires one forward pass of $D$ when using the projection discriminator~\cite{miyato2018cgans}.
The Algorithm~\ref{algo} presents the whole training procedure. We refer the readers to Appendix B for an illustration of Triple-GAN.

\section{Related Work}
\label{sec:related_work}

Various approaches have been developed to learn directed DGMs, 
including variational autoencoders (VAEs)~\cite{kingma2013auto,rezende2014stochastic},
generative moment matching networks (GMMNs)~\cite{li2015generative,dziugaite2015training},
generative adversarial nets (GANs)~\cite{goodfellow2014generative}
and 
autoregressive models~\cite{gregor2013deep,van2016conditional}.
These learning criteria are systematically compared in~\cite{theis2015note} and among them, GANs have proven effective on generating realistic samples~\cite{radford2015unsupervised,arjovsky2017wasserstein,karras2017progressive,karras2018style,brock2018large} and becomes popular in applications~\cite{isola2017image,zhu2017unpaired,zhang2017stackgan,ledig2017photo}.

Recent work has introduced inference networks in GANs for representation learning. In ALI~\cite{dumoulin2016adversarially,donahue2016adversarial}, the inference network approximates the posterior distribution of latent variables given true data in an unsupervised manner.
ALICE~\cite{li2017alice} enhances ALI by introducing conditional entropy regularization in the perspective of joint distribution matching. JointGAN~\cite{pu2018jointgan} extends ALI to multiple domains and achieves excellent results.
Triple-GAN also has an inference network (classifier) as in these work but there exist two important differences in the global equilibria and the objective functions between them:
(1) Triple-GAN matches both the distributions defined by the generator and classifier to true data distribution while ALI only ensures that the distributions defined by the generator and the inference network to be the same; (2) the discriminator will reject the samples from the classifier in Triple-GAN while the discriminator will accept the samples from the inference network in ALI, which leads to different update rules for the discriminator and inference network. These differences naturally arise because Triple-GAN is proposed to solve the problems of GANs with limited supervision as analyzed in Sec.~\ref{sec:introduction}. Indeed, ALI~\cite{dumoulin2016adversarially} uses the same approach as Improved-GAN~\cite{salimans2016improved} to deal with partially labeled data and hence it still suffers from the same problems (See results in Tab.~\ref{table:ssl_noda}).
Following ALI, InfoGAN~\cite{chen2016infogan} and Graphical-GAN~\cite{li2018graphical} disentangle explainable factors from other latent features in a purely unsupervised manner. However, such methods highly depend on proper inductive biases of the model and the data~\cite{locatello2018challenging}. Besides, the meaning of the disentangled factors cannot be predefined. In comparison, Triple-GAN can avoid the issues under the help of the given supervision. 

DGMs can be naturally extended to handle partially labeled data. For instance,
the conditional VAE~\cite{kingma2014semi} and its variants~\cite{maaloe2016auxiliary,li2017max} treat the missing labels as latent variables and infer them for the unlabeled data. 
Apart from the work mentioned in Sec.~\ref{sec:introduction}, BadGAN~\cite{dai2017good} is another semi-supervised GAN-based approach, which is based on the same observation that the objective functions in existing semi-supervised GANs are incompatible. However, BadGAN focuses on the classification problem and trains the generator to attack the decision boundary of the classifier instead of to match the data distribution. Therefore, BadGAN achieves comparable classification results with Triple-GAN but leave the generation problem unsolved. Triangle-GAN~\cite{gan2017triangle} extends our conference version~\cite{chongxuan2017triple} by minimizing the Jensen Shannon divergence between every two joint distributions and demonstrating the effectiveness in applications such as the image to image translation.

Despite the DGMs, there are extensive discriminative approaches for semi-supervised classification~\cite{laine2016temporal,dai2017good,luo2018smooth,oliver2018realistic,lee2013pseudo,miyato2015distributional,tarvainen2017mean, iscen2019label,athiwaratkun2018there}. Such methods define different regularization terms on the unlabeled data and often outperform generative approaches using a comparable classifier. Among them, mean teacher~\cite{tarvainen2017mean} is a competitive method and many recent methods~\cite{luo2018smooth,oliver2018realistic,iscen2019label,athiwaratkun2018there} are built upon it.
Our Triple-GAN-V2 also employs the mean teacher classifier~\cite{tarvainen2017mean} and we directly compare with it in Sec.~\ref{sec:experiments}.
We mention that there are some concurrent work~\cite{verma2019interpolation,sohn2020fixmatch,berthelot2019remixmatch,berthelot2019mixmatch} using more sophisticated classifiers based on the mixup training~\cite{zhang2017mixup} and fine-tuned data augmentation strategy. In principle, such methods can be incorporated in Triple-GAN to further improve the performance potentially, which is left as our future work.

\begin{table*}[t]
	\caption{Benchmark results (error rates \%) {\it without} data augmentation. The methods with $^\dagger$ use a similar {\it 13-layer CNN} classifier. The results with $^\mathsection$ are trained with more than 500,000 extra unlabeled data on SVHN.	All baseline results are from the corresponding references unless mentioned. As suggested by~\cite{oliver2018realistic}, we implement the most direct baselines~\cite{salimans2016improved,tarvainen2017mean} in our code base for a fair comparison. Our results are averaged by 3 runs.}
  \label{table:ssl_noda}
  \centering
  \begin{tabular}{lccccccc}
    \toprule
	Dataset  & SVHN  & SVHN & SVHN  & CIFAR10 & CIFAR10 & CIFAR10 & Tiny ImageNet \\
	Number of labels & $250$ & $500$ & $1,000$ & $1,000$ & $2,000$ & $4,000$ & $1,000$\\
    \midrule
    {\it M1+M2}~\cite{kingma2014semi} &  & & $36.02$ ($\pm0.10$) & & \\
    {\it Ladder}~\cite{rasmus2015semi} & & & & && 20.40 ($\pm0.47$)\\
    {\it ADGM}~\cite{maaloe2016auxiliary}$^\mathsection$  & & & 22.86 \\
    {\it SDGM}~\cite{maaloe2016auxiliary}$^\mathsection$  &  & & 16.61 ($\pm 0.24$) \\
    {\it MMCVA}~\cite{li2017max}$^\mathsection$  & & & 4.95 ($\pm0.18$)  \\
    {\it CatGAN}~\cite{springenberg2015unsupervised} & & & & & &  19.58 ($\pm0.58$) \\
    {\it ALI}~\cite{dumoulin2016adversarially}$^\dagger$& && 7.3 & & & 18.3 \\
    \midrule
    \textbf{-Using classifier in~\cite{salimans2016improved}-}\\
    ~~~~{\it Improved-GAN}~\cite{salimans2016improved}$^\dagger$ & &  &8.11 ($\pm1.3$)  & & & 18.63 ($\pm2.32$) &  28.47 ($\pm 0.41$)\\
    ~~~~{\it Triple-GAN-V1}  (\textbf{ours})$^\dagger$ & & & 5.77 ($\pm0.17$)& & & 16.99 ($\pm 0.36$) &  24.83 ($\pm 1.23$) \\
    \midrule
    {\it Triangle-GAN}~\cite{gan2017triangle}$^\dagger$ & & & & && 16.80 ($\pm 0.42$) \\
    {\it BadGAN}~\cite{dai2017good}$^\dagger$ & & & 4.25 ($\pm 0.03$)& &  & 14.41 ($\pm0.30$)\\
    {\it $\Pi$ model}~\cite{laine2016temporal}$^\dagger$ & 10.36 ($\pm 0.94$) & 7.01 ($\pm 0.29$) &  5.73 ($\pm 0.16$) & 32.18 ($\pm 1.33$)  & 23.92 ($\pm 1.07$) & 17.08 ($\pm 0.32$) \\
    {\it Pseudo Label}~\cite{oliver2018realistic}$^\dagger$ & & & 7.62 ($\pm 0.29$) & & & 17.78 $(\pm 0.57)$\\
    {\it SNTG}~\cite{luo2018smooth}$^\dagger$ & & & 4.02 ($\pm 0.20$) & && 12.49 ($\pm 0.36$) \\   
    {\it VAT}~\cite{miyato2015distributional}$^\dagger$ &  & & 5.77 $(\pm 0.32)$ & && 14.18 ($\pm 0.38$) \\
    {\it VAT+Ent}~\cite{miyato2015distributional}$^\dagger$ & & & 4.28 ($\pm 0.10$) & && 13.15 ($\pm 0.21$) \\
    {\it MT}~\cite{tarvainen2017mean}$^\dagger$ & 5.85 ($\pm 0.62$) & 5.45 ($\pm 0.14$) & 5.21 ($\pm 0.21$) &  30.62 ($\pm 1.13$) & 23.14 ($\pm 0.46$) & 17.74 ($\pm 0.30$) \\
    \midrule
    \textbf{-Using classifier in~\cite{tarvainen2017mean}-}\\
    ~~~~{\it MT} (our code base)$^\dagger$  & 5.82 ($\pm 0.87$) & 4.48 ($\pm 0.17$) &  5.39 ($\pm 0.12$)  & 28.34 ($\pm 0.36$) & 21.24 ($\pm 0.79$) & 16.37 ($\pm 0.21$) &  20.67 ($\pm 1.36$)\\
  ~~~~{\it Triple-GAN-V2} (\textbf{ours})$^\dagger$  & \textbf{4.19} ($\pm 0.62$) & \textbf{3.84} ($\pm 0.17$)  & \textbf{3.96} ($\pm 0.15$) &  \textbf{18.19} ($\pm 0.35$) &  \textbf{14.74} ($\pm 0.89$) &  \textbf{12.41} ($\pm 0.41$) & \textbf{17.40}  ($\pm 1.06$)\\
    \bottomrule
  \end{tabular}
\end{table*}

Some preliminary results were published in~\cite{chongxuan2017triple}. We extend the original version by presenting the framework in a more coherent and detailed way, extending it to 
the challenging extreme low data regime, proposing an enhanced version (i.e., Triple-GAN-V2), comparing to a large family of stronger classification (with or without standard data augmentation) and generation baselines,
and presenting new results on the Tiny ImageNet dataset.

\section{Experiments}
\label{sec:experiments}

In this section, we first present the datasets and the experimental settings briefly. Then, we show the classification and conditional generation results in both semi-supervised learning and the extreme low data regime.

\subsection{Datasets and Settings}

We evaluate Triple-GAN on the widely adopted MNIST~\cite{lecun1998gradient}, SVHN~\cite{netzer2011reading}, and CIFAR10~\cite{krizhevsky2009learning} and Tiny ImageNet datasets. MNIST consists of 50,000 training samples, 10,000 validation samples and 10,000 testing samples of handwritten digits of size $28\times 28$. SVHN consists of 73,257 training samples and 26,032 testing samples and each is a colored image of size $32\times 32$, containing a sequence of digits with various backgrounds. CIFAR10 consists of colored images distributed across 10 general classes---{\it airplane}, {\it automobile}, {\it bird}, {\it cat}, {\it deer}, {\it dog}, {\it frog}, {\it horse}, {\it ship} and {\it truck}. There are 50,000 training samples and 10,000 testing samples of size $32\times 32$ in CIFAR10. We split 5,000 training data of SVHN and CIFAR10 for validation if needed.
On CIFAR10, we follow~\cite{laine2016temporal} to perform ZCA for the input of $C$ but still generate and discriminate the raw images using $G$ and $D$, respectively. The Tiny ImageNet dataset consists of natural images of size $64 \times 64$. 
We select 10 classes, including {\it lion}, {\it snow mountain}, {\it coffee}, {\it teddy bear}, {\it penguin}, {\it cat}, {\it tower}, {\it butterfly}, {\it car} and {\it cabin}, out of the 200 classes and downscale the images to $32 \times 32$ (the dataset will be released with the source code). Each class consists of 500 training samples and 50 testing samples with diverse visual appearance. We randomly split ten percent of the training samples as the validation set if required.

We implement our method based on Theano~\cite{2016arXiv160502688short} and PyTorch~\cite{paszke2019pytorch}. We briefly summarize our experimental settings.\footnote{Our code is available at  https://github.com/zhenxuan00/triple-gan (Theano) and https://github.com/taufikxu/Triple-GAN (PyTorch).}
For a fair comparison, all the classification results of the baselines are from the corresponding papers or obtained based on the implementation released by the corresponding authors.
The generator and classifier of Triple-GAN-V1 and Triple-GAN-V2 have comparable architectures to those of the baselines~\cite{springenberg2015unsupervised,salimans2016improved,zhang2018dada,tarvainen2017mean,miyato2018spectral,miyato2018cgans} in the corresponding settings (See details in Appendix E). Besides, we average the classification results of Triple-GAN by multiple runs with different random initialization and splits of the training data and report the mean error rates with the standard deviations following existing work~\cite{salimans2016improved,zhang2018dada}.
In our experiments, we find that the training techniques for the original two-player GANs~\cite{denton2015deep,salimans2016improved,miyato2018spectral,miyato2018cgans} are sufficient to stabilize the optimization of Triple-GAN.

\subsection{Semi-Supervised Learning}
\label{sec:exp_ssl}

In semi-supervised learning, we first show that Triple-GAN can make predictions accurately and control the semantics of the generated samples as a unified model. Further, we will demonstrate that Triple-GAN can effectively utilize the large amount of unlabeled data and obtain superior or comparable classification and generation results to strong baselines.

\subsubsection{Classification}\label{sec:exp_cla}

\begin{table*}[t]
  \caption{
  Benchmark results (error rates \%)  {\it with} standard data augmentation using a similar {\it 13-layer CNN} classifier. All results are from the corresponding references unless specified. As suggested by~\cite{oliver2018realistic}, we implement~\cite{tarvainen2017mean} in our code base for a fair comparison. Our results are averaged by 3 runs.}
  \label{table:ssl_da}
  \centering
  \begin{tabular}{lccccccc}
    \toprule
    Dataset & SVHN & SVHN & SVHN & CIFAR10 & CIFAR10 & CIFAR10 &  Tiny ImageNet \\
	Number of labels & $250$ & $500$ & $1,000$ & $1,000$ & $2,000$ & $4,000$ & $1,000$\\
    \midrule
    {\it $\Pi$ model}\cite{laine2016temporal}&  9.69 ($\pm 0.92$)  &  6.83 ($\pm 0.66$) &  4.95 ($\pm 0.26$)   & 27.36 ($\pm 1.20$)   & 18.02 ($\pm 0.60$) & 13.20 ($\pm 0.27$)  \\
    {\it VAT}~\cite{miyato2015distributional}  & && 5.42 ($\pm 0.22$) & && 11.36 ($\pm 0.34$)\\
    {\it VAT+Ent}~\cite{miyato2015distributional} & &&
    3.86 ($\pm 0.11$) &
    10.55 ($\pm 0.05$) \\
    {\it SWA-480}~\cite{athiwaratkun2018there} & && & 17.48 ($\pm 0.13$) & 13.09 ($\pm 0.80$) & 10.30 ($\pm 0.21$)  \\
    {\it Fast-SWA-480}~\cite{athiwaratkun2018there} & && & 16.84 ($\pm 0.62$) & 12.24 ($\pm 0.31$) & \textbf{9.86 ($\pm 0.27$)}\\
    {\it SNTG}~\cite{luo2018smooth}&  4.29 ($\pm 0.23$) & 3.99 ($\pm 0.24$) & 3.86 ($\pm 0.27$) &  18.41 ($\pm 0.52$) & 13.64 ($\pm 0.32$) & 10.93 ($\pm 0.14$) \\
    {\it Label Propagation}~\cite{iscen2019label} & && & 16.93 ($\pm 0.70$)  & 13.22 ($\pm 0.29$) & 10.61 ($\pm 0.28$) \\
    {\it MT}~\cite{tarvainen2017mean} & 4.35 ($\pm 0.50$) &  4.18 ($\pm 0.27$) & 3.95 ($\pm 0.19$) &  21.55 ($\pm 1.48$) & 15.73 ($\pm 0.31$) & 12.31 ($\pm 0.28$)\\
    \midrule
    \textbf{-Using classifier in~\cite{tarvainen2017mean}-}\\
    ~~~~{\it MT} (our code base) & 4.89  ($\pm 0.28$) & 4.36 ($\pm 0.22$) & 4.48 ($\pm 0.17$) & 25.14 ($\pm 1.32$) & 18.05 ($\pm 0.45$) & 13.18 ($\pm 0.21$) & 19.07 ($\pm 0.42$)\\
    ~~~~{\it Triple-GAN-V2}   (\textbf{ours})  & \textbf{3.48} ($\pm 0.06$) & \textbf{3.61} ($\pm 0.26$) & \textbf{3.45} ($\pm 0.11$) & \textbf{15.00} ($\pm 0.57$) & \textbf{11.87} ($\pm 0.20$) & \textbf{10.01} ($\pm 0.16$) & \textbf{17.13}  ($\pm 1.40$) \\
    \bottomrule
  \end{tabular}
\end{table*}

\textbf{Results without Data Augmentation.} In Tab.~\ref{table:ssl_noda}, we compare our method with a large body of approaches without any data augmentation on the MNIST, SVHN, CIFAR10 and Tiny ImageNet datasets, respectively.
The results of Triple-GAN-V1 has been published in our conference version~\cite{chongxuan2017triple}. 
On all of the datasets, Triple-GAN-V1
outperforms the most direct competitor (i.e. Improved-GAN) substantially and consistently with the same classifier, which demonstrate the benefit of our compatible learning objective functions. 
In this paper, we add an enhanced Triple-GAN-V2 to compare against recent discriminative semi-supervised learning methods~\cite{laine2016temporal,dai2017good,luo2018smooth,oliver2018realistic,lee2013pseudo,miyato2015distributional,tarvainen2017mean, iscen2019label,athiwaratkun2018there}. 
Triple-GAN-V2 is built upon a mean teacher (MT) classifier~\cite{tarvainen2017mean}, and a projection discriminator~\cite{miyato2018cgans} with spectral normalization~\cite{miyato2018spectral}. Note that decoupling the hypothesis spaces of the classifier and discriminator as in Triple-GAN is necessary to adopt such advances simultaneously because they employ very different architectures and loss functions. In the last two rows of Tab.~\ref{table:ssl_noda}, we can see that Triple-GAN-V2 outperforms MT significantly, especially when the number of labeled data is extremely restricted. For instance, Triple-GAN-V2 reduces more than 10$\%$ error rate on CIFAR10 given 1,000 labels. This is because the pseudo discriminative loss can provide relatively more supervision in such cases.
Further, Triple-GAN-V2 is better than a large number of recent methods with comparable architectures, showing its effectiveness.

\begin{figure}[t]
\centering
\subfigure[SVHN]{\includegraphics[width=0.45\columnwidth]{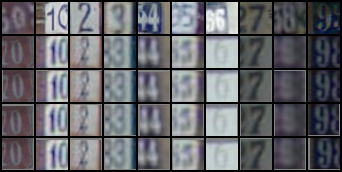}}
\subfigure[CIFAR10]{\includegraphics[width=0.45\columnwidth]{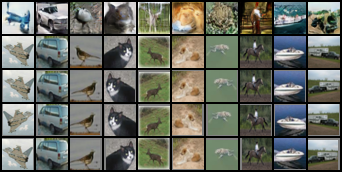}}
\vspace{-.1cm}
\caption{Comparison between the generated data and augmented data. The first row shows the data generated from Triple-GAN-V2. The second row shows the corresponding nearest neighbours in the training data of the same class. The last three rows show the randomly augmented data. The diversity of the generated data is higher than the augmented ones and the patterns of the two types of data are different.}
\label{fig:gen_vs_da}
\end{figure}

\textbf{Results with Data Augmentation.} Recent discriminative methods~\cite{luo2018smooth,oliver2018realistic,lee2013pseudo,miyato2015distributional,tarvainen2017mean, iscen2019label,athiwaratkun2018there} on semi-supervised learning often obtain a much better results with data augmentation. 
Therefore, we investigate whether Triple-GAN-V2 can still improve the MT baseline when standard data augmentation~\cite{tarvainen2017mean} is applied. The quantitative results are summarized in Tab.~\ref{table:ssl_da}. Though the gap between Triple-GAN-V2 and MT is slightly reduced compared to Tab.~\ref{table:ssl_noda}, the main conclusion that Triple-GAN-V2 can significantly outperform the baselines with comparable architectures still holds. Besides, Triple-GAN-V2 can benefit from the data augmentation technique as well. 
These results suggest that the generated data may have very different patterns from the randomly augmented ones and hence can provide complementary learning signals to the classifier. In addition, we qualitatively compare the generated data and the augmented ones in Fig.~\ref{fig:gen_vs_da}. These results strengthen the motivation of using GANs when data augmentation is applicable.

\begin{table}[t]
  \caption{Preliminary results on CIFAR-10 {\it with} standard data augmentation using a {\it 26-layer ResNet}~\cite{he2016deep} classifier with the ShakeShake regularization~\cite{gastaldi2017shake}.  All results are implemented in our code base and averaged by 3 runs.}
  \label{table:resnet}
  \centering
  \begin{tabular}{lccc}
    \toprule
	Number of labels & $1,000$ & $4,000$\\
    \midrule
    \textbf{-Using classifier in~\cite{tarvainen2017mean}-}\\
    ~~~~{\it MT} (our code base) &9.64 ($\pm 0.28
    $) &  6.61 ($\pm 0.17$) \\
    ~~~~{\it Triple-GAN-V2}   (\textbf{ours}) & \textbf{8.41} ($\pm 0.19$) & 6.54 ($\pm 0.08$) \\
    \bottomrule
  \end{tabular}
\end{table}

We provide preliminary results on CIFAR10 using a 26-layer ResNet classifier~\cite{tarvainen2017mean} in Tab.~\ref{table:resnet}. Triple-GAN-V2 obtains a significant improvement (compared to the measurement variance) over MT given 1,000 labels. The results of 4,000 labels indicates that the generator used in Triple-GAN-V2 cannot provide useful learning signals to a classifier with a 6.61$\%$ error rate. The result is reasonable and we believe it can be improved by using better GANs. We also notice that some concurrent work~\cite{verma2019interpolation,sohn2020fixmatch,berthelot2019remixmatch,berthelot2019mixmatch} adopts the mixup training~\cite{zhang2017mixup} based on the 26-layer ResNet classifier and fine-tuned data augmentation strategies. In principle, such methods can be incorporated in Triple-GAN to further improve the performance potentially, which is left as our future work. 

\begin{table}[t]
	\caption{Ablation study of the main factors in Triple-GAN-V2. MT refers to the mean teacher classifier~\cite{tarvainen2017mean} and ER refers to the entropy regularized classifier~\cite{yves2004semi}. For fairness, we implement all methods in our code base.}
  \label{table:ablation}
  \centering
  \begin{tabular}{cccccc}
    \toprule
Method &  $\mathcal{R_U}$ & $\alpha_{\mP}$ & Augmentation & Error rate (\%) \\
   \midrule
Triple-GAN &  MT & 10.0  &  
  \cmark & 11.97 \\
Triple-GAN &   MT & 1.0  &  \cmark & 10.90  \\
Triple-GAN & MT & 0.0 &   \cmark & 13.68 \\
\midrule
Triple-GAN &    MT & 10.0 &  \xmark & 13.78 \\
Triple-GAN &  MT & 1.0 &  \xmark &  13.10 \\
Triple-GAN & MT & 0.0 &    \xmark &  15.90 \\
 \midrule
Triple-GAN & MT & 3.0  &  \cmark & 10.01 \\
Baseline &  MT & - & \cmark& 13.18 \\
\midrule
Triple-GAN &   MT & 3.0 &  \xmark & 12.41 \\
Baseline &    MT &  - & \xmark& 16.37\\
 \midrule
Triple-GAN &   ER & 3.0  &   \cmark & 12.98 \\
Baseline &  ER & - & \cmark& 18.23 \\
\midrule
Triple-GAN &   ER & 3.0  &   \xmark & 15.44 \\
Baseline &  ER &  -  & \xmark& 21.98 \\
   \midrule
Triple-GAN &   \xmark &  3.0  &  \cmark & 13.91\\
Baseline &  \xmark & - & \cmark& 21.48 \\
\midrule
Triple-GAN &   \xmark &  3.0  &  \xmark & 17.88\\
Baseline &  \xmark & -  & \xmark& 24.21\\
   \bottomrule
  \end{tabular}
\end{table}

\textbf{Ablation Study.} We provide a systematical ablation study of Triple-GAN-V2 in Tab.~\ref{table:ablation}. Firstly, we investigate the effect of $\alpha_{\mP}$. Simply setting $\alpha_{\mP} = 0$ will lead to a significant drop of accuracy, indicating the importance of the pseudo discriminative loss in Triple-GAN. The result of $\alpha_{\mP} = 3.0$ is the best while those of other values still outperform the baseline significantly. Secondly, we change the unlabeled data loss and the data augmentation strategy. We can see that no matter which loss is used and whether the data augmentation is applied or not, Triple-GAN-V2
is superior to the corresponding baseline. We provide more results on the ablation study and the convergence speed of Triple-GAN-V1 in Appendix C.

\subsubsection{Generation}\label{sec:exp_gen}

\begin{table}[t]
	\caption{IS~\cite{salimans2016improved} and FID~\cite{heusel2017gans} on CIFAR10. Results with $^\mathsection$, $^\dagger$  $^{\ddagger}$ and $^{\dagger\dagger}$ are trained with fully unlabeled data,
	partially labeled data (4,000 labels), fully labeled data, and a small subset of fully labeled data (4,000 samples), respectively. The baseline results are from the corresponding references unless specified. For a fair comparison, we use the same protocol and Inception models as in CGAN-PD~\cite{miyato2018cgans} for evaluation.}
  \label{table:is_fid}
  \centering
  \begin{tabular}{lcccc}
    \toprule
Metrics  & IS $\uparrow$ & FID $\downarrow$ \\
    \midrule
    \textbf{-Using GAN in~\cite{salimans2016improved}-} \\
    ~~~~{\it Improved-GAN-FM}~\cite{salimans2016improved}$^\dagger$ & 3.87 \\
    ~~~~{\it Triple-GAN-V1}~\cite{chongxuan2017triple} (\textbf{ours})$^\dagger$  & 5.08 ($\pm0.09$) \\
    \midrule
    \textbf{-Using ResNet GAN-} \\
    ~~~~{\it DCGAN}~\cite{radford2015unsupervised}$^\mathsection$ & 6.16 $(\pm 0.07)$~\cite{gulrajani2017improved} & 36.9~\cite{heusel2017gans} \\
    ~~~~{\it DCGAN}~\cite{radford2015unsupervised}$^{\ddagger}$ & 6.58~\cite{gulrajani2017improved} &  \\
    ~~~~{\it WGAN-GP}~\cite{gulrajani2017improved}$^\mathsection$ & 7.86 $(\pm 0.07)$ & 24.8~\cite{heusel2017gans}\\
    ~~~~{\it WGAN-GP}~\cite{gulrajani2017improved}$^{\ddagger}$ & 8.42 $(\pm 0.10)$\\
    ~~~~{\it ACGAN}~\cite{odena2016conditional}$^{\ddagger}$ & 8.22~\cite{miyato2018cgans} & 19.7~\cite{miyato2018cgans}\\
    \midrule
    \textbf{-Using GAN in~\cite{miyato2018cgans}-} \\
    ~~~~{\it SNGAN}~\cite{miyato2018spectral}$^\mathsection$ & 8.22 ($\pm 0.05$) & 21.7 ($\pm 0.21$) \\
    ~~~~{\it CGAN-PD}~\cite{miyato2018cgans}$^{\ddagger}$  & \textbf{8.62} & \textbf{17.5} \\
    ~~~~{\it CGAN-PD} (our code base)$^{\dagger\dagger}$ & 6.03 & 42.3\\
   ~~~~{\it Triple-GAN-V2}   (\textbf{ours})$^\dagger$ & 8.51 & 17.9 \\
    \bottomrule
  \end{tabular}
\end{table}

We demonstrate that Triple-GAN can learn good $G$ and $C$ simultaneously by quantitatively and qualitatively evaluating the samples generated from the exact models used in~Sec.~\ref{sec:exp_cla}. For a fair comparison, the generative model and the number of labels are the same as the corresponding baselines~\cite{salimans2016improved,miyato2018spectral,miyato2018cgans}.

\textbf{Quantitative Results.} In Tab.~\ref{table:is_fid}, we quantitatively compare to existing methods in terms of the Inception Score (IS)~\cite{salimans2016improved} and Fr\'echet Inception Distance (FID)~\cite{heusel2017gans} on the widely adopted CIFAR10 dataset. Firstly, we note that Triple-GAN-V2 trained with only 8$\%$ labels achieves a comparable result to the strongest baseline (i.e., CGAN with a projection discriminator ({\it CGAN-PD})~\cite{miyato2018cgans}) trained with full labels. Secondly, we can see that Triple-GAN-V2 outperforms the CGAN-PD trained on 4,000 labeled data and the spectral normalized GAN ({\it SNGAN})~\cite{miyato2018spectral} trained on 50,000 unlabeled data substantially. Such results suggest that Triple-GAN can exploit the large amount of unlabeled data and the small amount of labels to generate high quality samples, which strongly motivates Triple-GAN given partially labeled data.

\begin{figure}[t]
\centering
\subfigure[MNIST data]{\includegraphics[height=0.45\columnwidth]{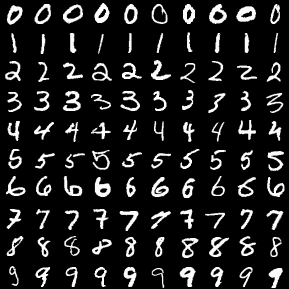}}
\subfigure[MNIST samples]{\includegraphics[width=0.45\columnwidth]{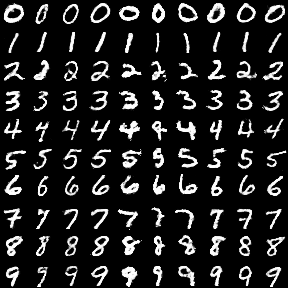}}
\subfigure[SVHN data]{\includegraphics[height=0.45\columnwidth]{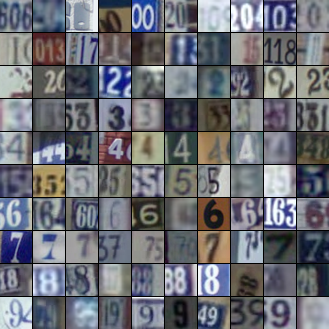}}
\subfigure[SVHN samples]{\includegraphics[width=0.45\columnwidth]{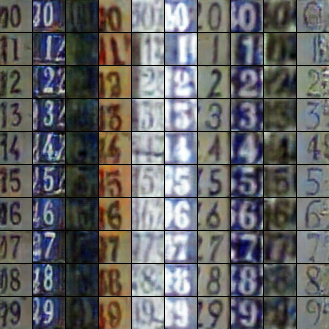}}
\subfigure[CIFAR10 data]{\includegraphics[width=0.45\columnwidth]{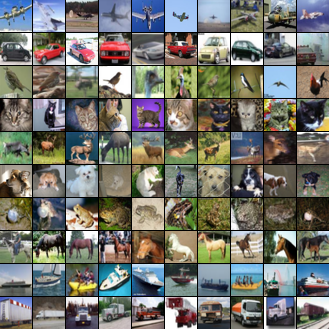}}
\subfigure[CIFAR10 samples]{\includegraphics[height=0.45\columnwidth]{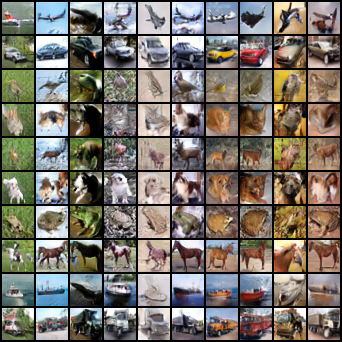}}
\subfigure[Tiny Imagenet data]{\includegraphics[width=0.45\columnwidth]{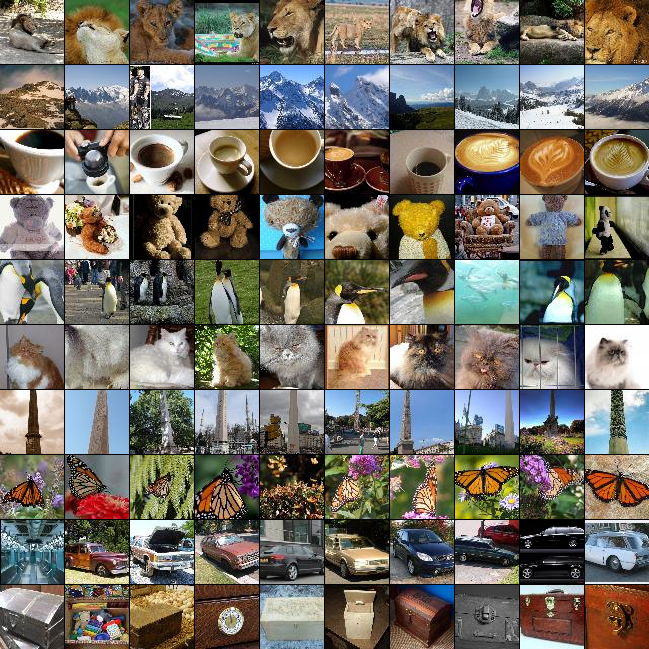}}
\subfigure[Tiny Imagenet samples]{\includegraphics[height=0.45\columnwidth]{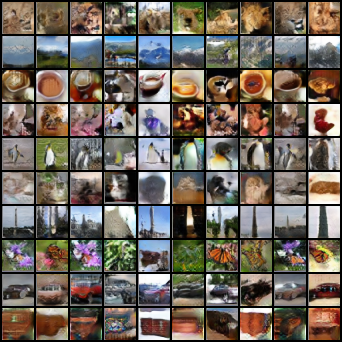}}
\vspace{-.1cm}
\caption{Samples of Triple-GAN in semi-supervised learning. On MNIST and SVHN, the samples are from Triple-GAN-V1. On CIFAR10 and Tiny ImageNet, the samples are from Triple-GAN-V2. In the figures, each row shares the same label and each column shares the same latent variables.}
\vspace{-.1cm}
\label{fig:dis}
\end{figure}

\textbf{Qualitative Results.} We show the ability of Triple-GAN to disentangle category and latent style features in Fig.~\ref{fig:dis}. It can be seen that most of the samples have correct semantics and the latent features in Triple-GAN encode meaningful physical factors, including the scale, intensity, orientation, color and so on.
Some GANs~\cite{radford2015unsupervised,dumoulin2016adversarially,odena2016conditional} can generate data class-conditionally given full labels,
while Triple-GAN can do the same thing given much less label information. 
We provide more conditional sample results and latent space interpolation results in Appendix D.


\subsection{The Extreme Low Data Regime}

We evaluate Triple-GAN on a small subset of the SVHN, CIFAR10 and Tiny ImageNet datasets to show the ability of classification and generation in the extreme low data regime. We remove the losses on the unlabeled data as presented in Sec.~\ref{sec:trick} and keep all the networks the same as those in semi-supervised learning.

\subsubsection{Classification}

\begin{table}[t]
	\caption{Preliminary results (error rates \%)  using a similar {\it 13-layer CNN} classifier. The results with $^\dagger$ are from the corresponding references. The classifiers in methods with $^\mathsection$ are trained {\it with} standard data augmentation and the ones in others are trained {\it without} any data augmentation. As suggested by~\cite{oliver2018realistic}, we implement the CNN and MT~\cite{tarvainen2017mean} baselines in our code base for a fair comparison. Our results are averaged by 3 runs.}
  \label{table:low}
  \centering
  \resizebox{\columnwidth}{21mm}{
  \begin{tabular}{lcccccc}
    \toprule
	Datasets &SVHN & CIFAR10 & Tiny  \\
	 && & ImageNet\\
	Number of labels  & $1,000$ & $4,000$ & $2,000$\\
    \midrule
    {\it Improved-GAN}~\cite{salimans2016improved} & 25.47$^{\dagger}$\\
    {\it DADA}~\cite{zhang2018dada} &  24.17$^{\dagger}$ & 33.57 ($\pm 0.46$) & 29.73 ($\pm 2.02$) \\
    {\it CNN} (our code base) & 22.72 ($\pm 2.39$) & 46.10 ($\pm 3.18$) & 24.92 ($\pm 1.12$)\\
    {\it Triple-GAN-V1} (\textbf{ours})& 21.44 ($\pm 1.95$) & 32.73 ($\pm 0.86$) & 20.67 ($\pm 1.84$) \\
    \midrule
    {\it MT} (our code base) & 11.68 ($\pm 0.36$)  & 21.02 ($\pm 0.49$) &  17.00 ($\pm 0.53$) \\
    {\it Triple-GAN-V2}   (\textbf{ours}) & \textbf{11.08} ($\pm 0.27$) & \textbf{20.07} ($\pm 0.06$) & \textbf{16.27} ($\pm 1.29$)\\
    \midrule
    {\it MT} (our code base)$^\mathsection$ & 11.08 ($\pm 0.40$)  & 18.10 ($\pm 0.43$)& 15.93 ($\pm 1.67$) \\
    {\it Triple-GAN-V2}   (\textbf{ours})$^\mathsection$  & \textbf{10.17} ($\pm 0.60$)  & \textbf{17.34} ($\pm 0.26$) &  \textbf{15.40} ($\pm 0.40$)\\
    \bottomrule
  \end{tabular}
  }
\end{table}

We compare Triple-GAN-V1 with two baselines. 
The main competitor to Triple-GAN-V1 is DADA~\cite{zhang2018dada}, which is a conditional variant of Improved-GAN for the extreme low data regime setting. The {\it CNN} baseline denotes a classifier that shares the same architecture as that in Triple-GAN-V1 but is trained without any deep generative models. We compare Triple-GAN-V2 to the mean teacher ({\it MT}) classifier~\cite{tarvainen2017mean} in cases with and without data augmentation. 

Tab.~\ref{table:low} show the quantitative results on the SVHN, CIFAR10 and Tiny ImageNet datasets. 
Both Triple-GAN-V1 and Triple-GAN-V2 achieves better averaged results than the corresponding baselines. We note that the improvement of Triple-GAN is smaller than that in semi-supervised learning and sometimes is not significant, mainly because it is hard to generate high quality images given such a limited amount of data (See Fig.~\ref{fig:ld}). A potential improvement of Triple-GAN is to adopt more recent GAN training techniques such as~\cite{karras2020training}.
Nevertheless, in many cases, Triple-GAN outperforms the baselines significantly compared to the measurement variance and is at least comparable to the baselines in the other cases.


\subsubsection{Generation}

\begin{figure}[t]
\centering
\subfigure[CIFAR10 samples]{\includegraphics[width=0.45\columnwidth]{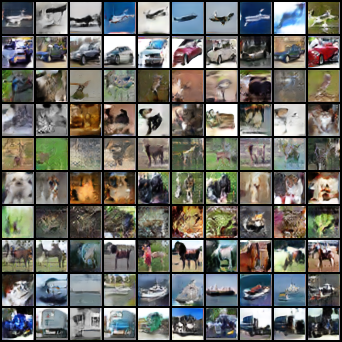}}
\subfigure[Tiny ImageNet samples]{\includegraphics[width=0.45\columnwidth]{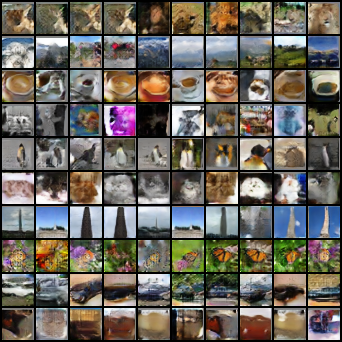}}
\caption{Conditional samples of Triple-GAN-V2 in the extreme low data regime. In the figures, Each row shares the same label.}
\label{fig:ld}
\end{figure}

Fig.~\ref{fig:ld} presents the conditional samples generated from Triple-GAN-V2 in the extreme low data regime on the CIFAR10 and Tiny ImageNet datasets. Compared to the results in semi-supervised learning (Fig.~\ref{fig:dis}), the quality and the diversity of the samples are worse because there is no unlabeled data available in the extreme low data regime. 
In fact,
the FID of Triple-GAN-V2 on CIFAR10 (4,000 samples) is $42.3$\footnote{This is the same as the result of CGAN-PD$^{\dagger\dagger}$~\cite{miyato2018cgans} in Tab.~\ref{table:is_fid} because no additional unlabeled data is available.}, which is significantly worse than $17.9$ in semi-supervised learning.
Nevertheless, Triple-GAN-V2 is still able to generate meaningful samples with correct semantics in most of the cases.

In summary, the extreme low data regime setting is much more challenging than semi-supervised learning according to our preliminary experiments. We hope the promising results of Triple-GAN can motivate future research in this setting.

\section{Conclusion}
\label{sec:conclusion}

We present Triple-GAN, a unified framework for classification and class-conditional generation in both semi-supervised learning and the extreme low data regime. Triple-GAN is formulated as a three-player minimax game between a classifier, a generator and a discriminator. Thanks to the theoretical insights, Triple-GAN is empirically effective and flexible to incorporate new techniques.
The enhanced Triple-GAN-V2 proposed in this paper significantly outperforms the original Triple-GAN-V1, showing the promise to further boost the performance by adopting the future advances in both semi-supervised classification and GANs. Following the baselines, we focus on images of size $32\times32$ in this paper and leave the extension on larger datasets like real ImageNet as our future work.


%

\ifCLASSOPTIONcompsoc
  \section*{Acknowledgments}
\else
  \section*{Acknowledgment}
\fi

This work was supported by the National Key Research and Development Program of China (No. 2017YFA0700904), NSFC Projects (Nos. 61620106010, 61621136008), Beijing NSF Project (No. L172037), Beijing Academy of Artificial Intelligence (BAAI), Tiangong Institute for Intelligent Computing, the JP Morgan Faculty Research Program and the NVIDIA NVAIL Program with GPU/DGX Acceleration. C. Li was supported by the Chinese postdoctoral innovative talent support program and Shuimu Tsinghua Scholar Program.

\ifCLASSOPTIONcaptionsoff
  \newpage
\fi




\bibliographystyle{IEEEtran}
\bibliography{IEEEexample.bib}

\appendices

\section{Detailed Theoretical Analysis}

\begin{lemma1}
For any fixed $C$ and $G$, the optimal discriminator $D$ of the game defined by the objective function $U(C,G,D)$ is
\begin{equation}\label{eq:opt_d}
D_{C, G}^*(x, y) = \frac{p(x, y)}{p(x, y) +
p_{\alpha} (x, y)},
\end{equation}
where $p_{\alpha} (x, y) := (1-\alpha)p_g(x, y) + \alpha p_c(x, y)$ is a mixture distribution for $\alpha \in (0, 1)$.
\end{lemma1}

\begin{proof}
Given the classifier and generator, the objective function can be rewritten as
\setlength{\arraycolsep}{0.0em}
\begin{eqnarray}\label{eq:proof_opt_d}
 &U&(C,G,D) \nonumber  \\
 &=& \iint  p(x, y) \log D(x, y)  dy dx \nonumber  \\
 &+& (1-\alpha) \iint p(y) p_z(z) \log (1 - D(G(z, y), y)) dy dz \nonumber \\
 &+& \alpha \iint p(x) p_c(y|x) \log (1 - D(x, y))  dy dx  \nonumber \\
 &= &\iint p(x, y) \log D(x, y) dy dx \nonumber \\
 &+&  \iint p_{\alpha} (x, y)\log (1 - D(x, y)) dy dx, \nonumber
\end{eqnarray}
which achieves the maximum at $\frac{p(x, y)}{p(x, y) + p_{\alpha}(x, y)} $.

\end{proof}

\begin{lemma2}
The global minimum of $V(C, G)$ is achieved if and only if $p(x, y) = p_{\alpha}(x, y) $.
\end{lemma2}
\begin{proof}

Given $D_{C, G}^*$, we can reformulate the minimax game with value function $U$ as:
\setlength{\arraycolsep}{0.0em}
\begin{eqnarray}\label{eq:u_cg}
V(C,G) =\iint p(x, y) \log \frac{p(x, y)}{p(x, y) + p_{\alpha}(x, y)} dy dx \nonumber \\
+ \iint p_{\alpha}(x, y) \log \frac{p_{\alpha} (x, y)}{p(x, y) +  p_{\alpha}(x, y)} dy dx  . \nonumber
\end{eqnarray}

Following the proof in GAN, the $V(C,G)$ can be rewritten  as
\setlength{\arraycolsep}{0.0em}
\begin{eqnarray}\label{eq:optimal}
V(C,G) = -\log 4 + 2 \mathbb{D}_{JS} (p(x, y) ||p_{\alpha}(x, y)),
\end{eqnarray}
where $\mathbb{D}_{JS}$ is the Jensen-Shannon divergence, which is always non-negative and the unique optimum is achieved if and only if $p(x, y) = p_{\alpha} (x, y)=  (1-\alpha)p_g(x, y) + \alpha p_c(x, y)$.
\end{proof}
\begin{corollary2}
Given $p(x, y) = p_{\alpha} (x, y)$, the marginal distributions of $p(x, y)$, $p_c(x, y)$ and $p_g(x, y)$ are the same, i.e. $p(x) = p_g(x) = p_c(x)$ and $p(y) = p_g(y) = p_c(y)$.
\end{corollary2}

\begin{proof}
Remember that $p_g(x, y) = p(y) p_g(x | y)$ and $p_c(x, y) = p(x) p_c(y | x)$. Take integral with respect to $x$ on both sides of $p(x, y) = p_{\alpha} (x, y)$ to get
$$
\int p(x , y) dx = (1-\alpha) \int p_g(x, y) dx + \alpha \int  p_c(x, y) dx,
$$
which indicates that
$$
p(y) = (1-\alpha) p(y) + \alpha p_c(y), \textrm{ i.e. } p_c(y) = p(y) = p_g(y).
$$
Similarly, it can be shown that $p_g(x) = p(x) = p_c(x)$ by taking integral with respect to $y$.
\end{proof}

\begin{theorem3}
The equilibrium of $ \tilde U(C, G, D)$ is achieved if and only if $p(x, y) = p_g(x, y) = p_c(x, y)$.
\end{theorem3}

\begin{proof}
According to the definition, we have $ \tilde U (C, G, D) = U(C,G,D) + \mR_{\mC} + \alpha_{\mP} \mP_{\mP}$, where 
$$
\mR_{\mC} = \ep_{p(x, y)}[- \log p_c(y | x)],
$$
and 
$$
\mP_{\mP} = \ep_{p_g(x, y)}[- \log p_c(y | x)].
$$

We can rewrite $\mR_{\mC}$ as:
\setlength{\arraycolsep}{0.0em}
\begin{eqnarray}\label{eq:proof_kl}
 \mR_{\mC} &=& \ep_{p(x, y)}[- \log p_c(y | x)] \nonumber\\
 &=& \ep_{p(x, y)}[ \log \frac{p(x, y)}{p_c(x, y)} ] - \ep_{p(x, y)}[ \log p(y|x) ] \nonumber\\ 
  &=& \mD (p(x, y)||p_c(x, y)) + \ep_{p(x)}\left [\mH[p(y|x)]\right ], \nonumber
\end{eqnarray}
where $\mD(\cdot || \cdot)$ denotes the KL-divergence and $\mH[\cdot]$ denotes the differential entropy. Because the second term is determined by the data distribution, minimizing $\mR_{\mC}$ is equivalent to minimizing $
\mD (p(x, y)||p_c(x, y))$. We now prove the equivalence of minimizing the pseudo discriminative loss $\mR_{\mP}$  and the KL-divergence $\mD (p_g(x, y)||p_c(x, y)) $ as follows:
\setlength{\arraycolsep}{0.0em}
\begin{eqnarray}
&\mD& (p_g(x, y)||p_c(x, y)) + \ep_{p_g(x)}\left[\mH[p_g(y|x)]\right] - \mD(p_g(x) || p(x)) \nonumber \\
&=& \iint p_g(x, y) \log \frac{p_g(x, y)}{p_c(x, y)} + p_g(x, y) \log \frac{1}{p_g(y|x)}dxdy \nonumber\\
&-& \int p_g(x) \log \frac{p_g(x)}{p(x)} dx \nonumber \\
&=& \iint p_g(x, y) \log \frac{p_g(x, y)p(x)}{p_c(x, y)p_g(y|x)p_g(x)}dxdy \nonumber \\
&=&  \ep_{p_g(x, y)}[-\log p_c(y | x)] = \mR_{\mP}. \nonumber
\end{eqnarray}
Note that $\ep_{p_g(x)}\left[\mH[p_g(y|x)]\right] - \mD(p_g(x) || p(x))$ is a constant with respective to $\theta_c$.
Therefore, if we only optimize the parameters in $C$, these two losses are equivalent. 

The two KL divergences are always non-negative and zero if and only if $p(x, y) = p_c(x , y)$ and $p_g(x, y) = p_c(x , y)$, respectively. 
Note that the previous lemmas can also be applied to $\tilde U (C, G, D)$, which indicates that $p(x, y) = p_{\alpha}(x, y)$ at the global equilibrium and concludes the proof.
\end{proof}

\begin{corollary3}
Adding any divergence (e.g. the KL divergence) between any two of the joint distributions or the conditional distributions or the marginal distributions to $\tilde U$
as an additional regularization to be minimized, will not change the global equilibrium of $\tilde U$.
\end{corollary3}
\begin{proof}
This conclusion can be straightforwardly obtained by the global equilibrium point of $\tilde U$ and the definition of a statistical divergence between two distributions.
\end{proof}

\begin{figure*}[t]
\centering
\includegraphics[width=1.98\columnwidth]{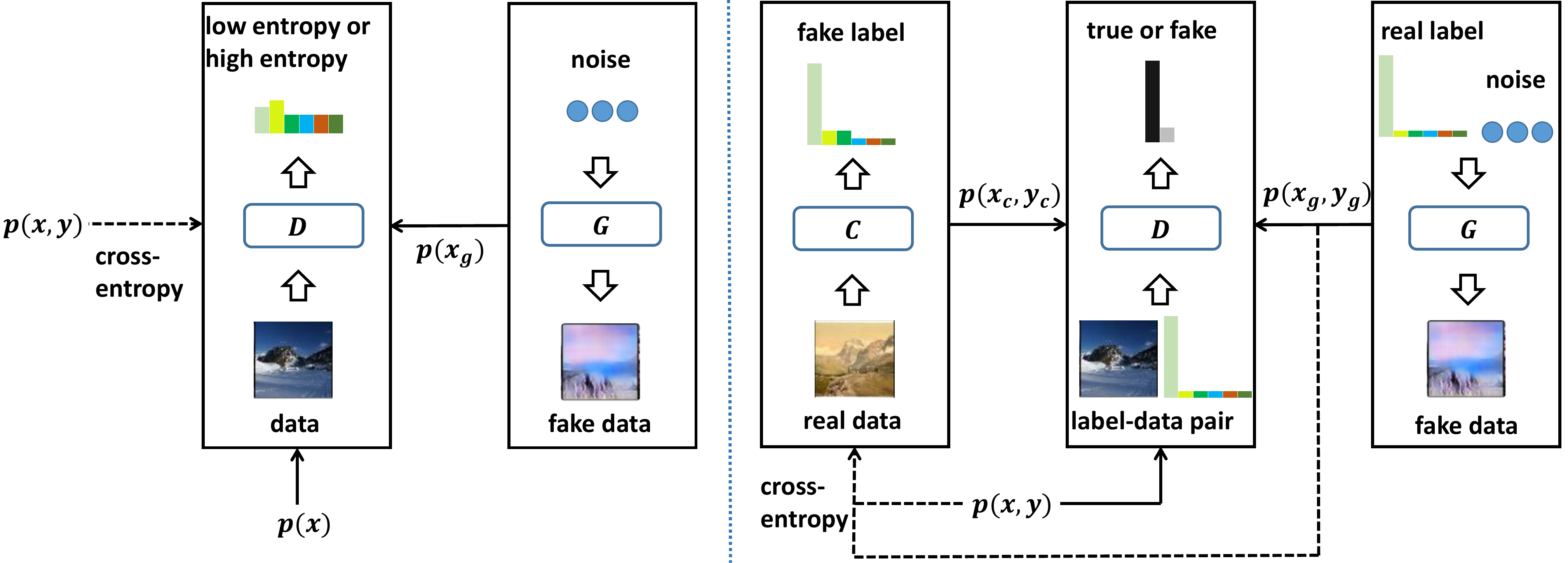}
\caption{Comparison between Triple-GAN (right panel) and a  representative prior work (left panel), i.e. CatGAN~\cite{springenberg2015unsupervised}. 
In CatGAN, the discriminator not only classifies the input but also estimates whether the input is real or fake. Besides, CatGAN focuses on the marginal distribution of images without considering label information. In Triple-GAN, the generator and the classifier approximate two conditional distributions respectively and the discriminator solely identifies the label-data pairs. Triple-GAN ensures that $p_c(x, y) = p_g(x, y) = p(x, y)$ after convergence.}
\label{fig:triple-gan-arc}
\end{figure*}

\section{Illustration of Triple-GAN}

We illustrate the training protocol of Triple-GAN in Fig.~\ref{fig:triple-gan-arc}.

\begin{figure}
\centering
\includegraphics[width=0.9\columnwidth]{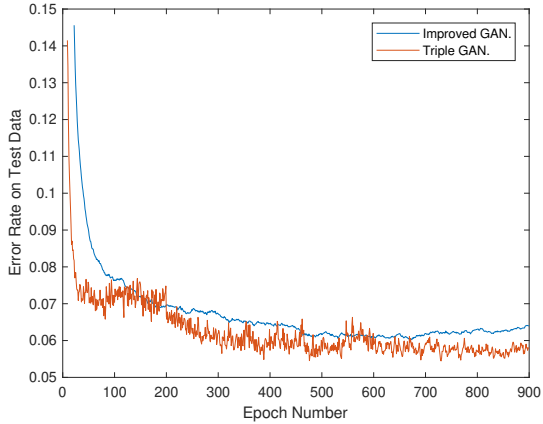}
\caption{The convergence speeds of Improved-GAN and Triple-GAN-V1 on SVHN.}
\label{fig:convergence}
\end{figure}

\section{More Classification Results of Triple-GAN-V1 in Semi-supervised Learning}

\textbf{Convergence Speed.} Though Triple-GAN-V1 has one more network, its convergence speed is at least comparable to that of Improved-GAN, as presented in Fig.~\ref{fig:convergence}. Both the models are trained on the SVHN dataset with the default settings and Triple-GAN-V1 can get good results in tens of epochs. The reason that the learning curve of Triple-GAN-V1 is oscillatory may be the larger variance of the gradients due to the presence of discrete variables. Also note that we apply the pseudo discriminative loss at the 200-th epoch and then the test error is reduced significantly in next 100 epochs.

\textbf{Ablation Study.} We investigate the reasons for the outstanding performance of Triple-GAN-V1. 
we focus on the comparison with Improved-GAN~\cite{salimans2016improved} because we use the exact same classifier, data split and code base, to ensure fairness as suggested by~\cite{oliver2018realistic}.
We train a single semi-supervised classifier without $G$ and $D$ on SVHN as the baseline and get a result of more than $10\%$ error rate, which shows that $G$ is important for semi-supervised learning even though the classifier can leverage unlabeled data directly. On CIFAR10, the baseline (a simple version of $\Pi$ model~\cite{laine2016temporal}) achieves 17.7\% error rate. The smaller improvement is reasonable as CIFAR10 consists of complex natural images and hence $G$ is not as good as that on SVHN.
In addition, we evaluate Triple-GAN-V1 without the pseudo discriminative loss on SVHN and it achieves about $7.8\%$ error rate, which shows the advantages of the compatible objective functions (better than the $8.11\%$ error rate of Improved-GAN) and the importance of the pseudo discriminative loss (worse than the complete Triple-GAN-V1 by $2\%$).

\begin{figure}[t]
\centering
\includegraphics[width=.95\columnwidth]{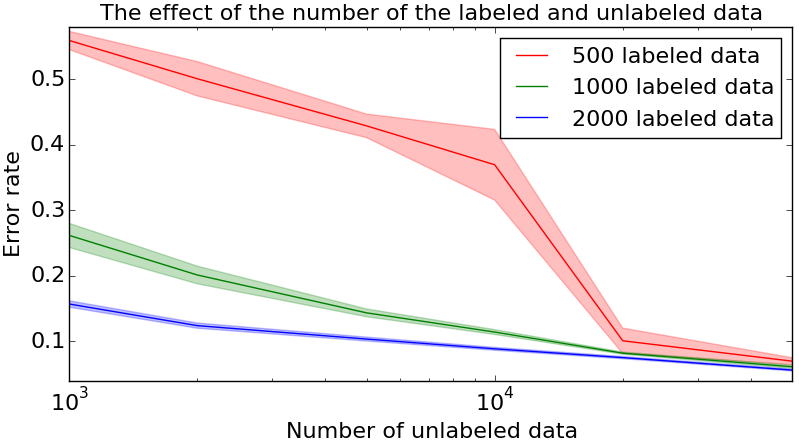}
\caption{Error rates ($\%$) on the partially labeled SVHN dataset with different numbers of labeled and unlabeled data. The results are averaged by 9 runs with different random selections of labeled and unlabeled data. We plot the mean and error bars of one standard deviation in each curve.}
\label{fig:label-unlabel}
\end{figure}

\textbf{Effect of the numbers of labeled and unlabeled data.} We systematically analyze the effect of the numbers of labeled and unlabeled data on the SVHN dataset in Fig~\ref{fig:label-unlabel}. As the number of labels increases, the performance of Triple-GAN-V1 is better and has a lower variance. However, if given a sufficient number of unlabeled data, i.e. more than 20,000, then the performance of Triple-GAN-V1 won't get hurt too much by reducing the number of labels from 2,000 to 500. The results suggest the importance of unlabeled data to the final performance of the classifier in semi-supervised learning.

\section{More Generation Results of Triple-GAN-V1 in Semi-supervised Learning}

\begin{figure}[t]
\centering
\subfigure[Feature Matching]{\includegraphics[width=0.45\columnwidth]{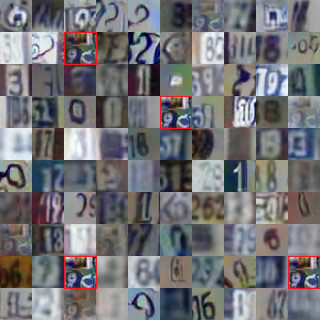}}
\subfigure[Triple-GAN]{\includegraphics[width=0.45\columnwidth]{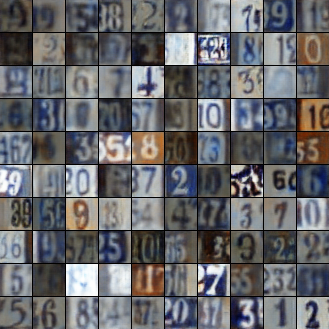}}
\caption{Comparison between samples from Improved-GAN trained with feature matching and Triple-GAN-V1.}
\label{fig:compare}
\end{figure}

\textbf{Sample Quality Comparison to Improved-GAN.} In Fig.~\ref{fig:compare}, we first compare the quality of the samples generated by Triple-GAN-V1 and Improved-GAN trained with feature matching~\cite{salimans2016improved},\footnote{Though the Improved-GAN trained with minibatch discrimination~\cite{salimans2016improved} can generate good samples, it fails to predict labels accurately.} which works well for semi-supervised classification.
We can see that Triple-GAN-V1 outperforms the baseline by generating fewer meaningless samples and clearer digits. Further, 
the baseline generates the same strange sample four times, labeled with the red rectangles in Fig.~\ref{fig:compare}.

\begin{figure}[t]
\centering
\subfigure[Automobile]{\includegraphics[width=0.45\columnwidth]{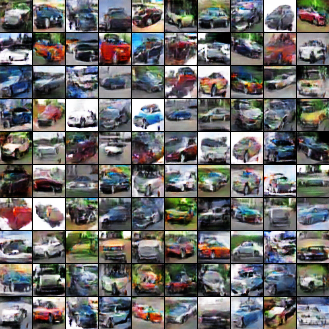}}
\subfigure[Horse]{\includegraphics[width=0.45\columnwidth]{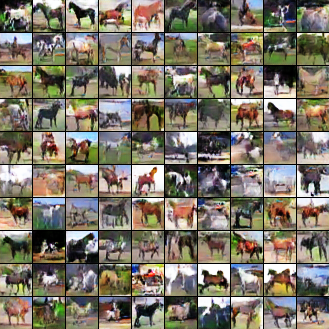}}
\subfigure[Bird]{\includegraphics[width=0.45\columnwidth]{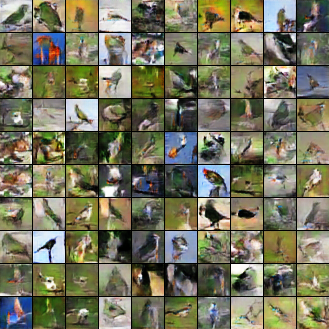}}
\subfigure[Ship]{\includegraphics[width=0.45\columnwidth]{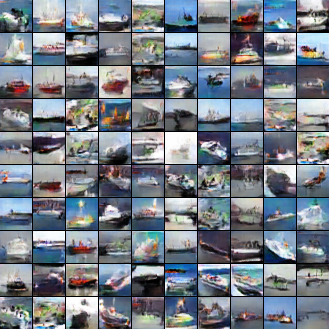}}
\subfigure[Airplane]{\includegraphics[width=0.45\columnwidth]{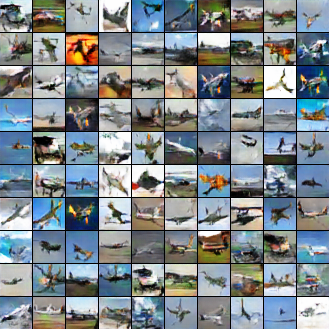}}
\subfigure[Cat]{\includegraphics[width=0.45\columnwidth]{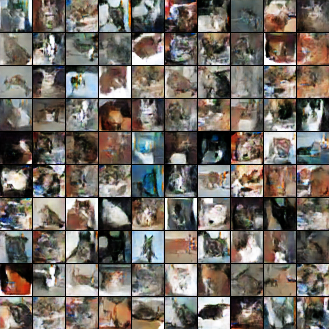}}
\subfigure[Deer]{\includegraphics[width=0.45\columnwidth]{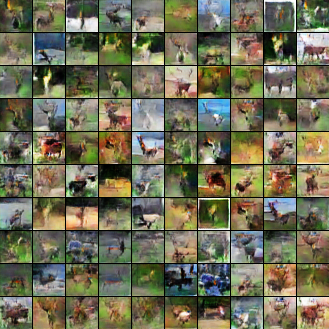}}
\subfigure[Dog]{\includegraphics[width=0.45\columnwidth]{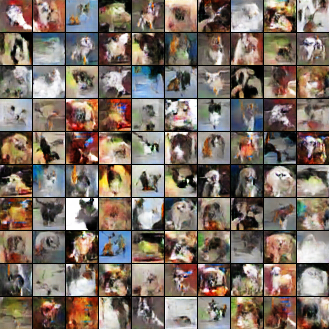}}
\subfigure[Frog]{\includegraphics[width=0.45\columnwidth]{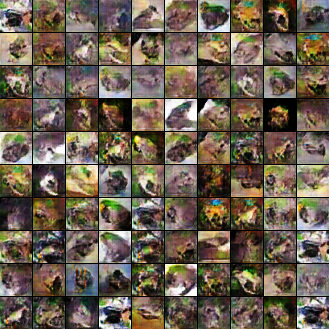}}
\subfigure[Truck]{\includegraphics[width=0.45\columnwidth]{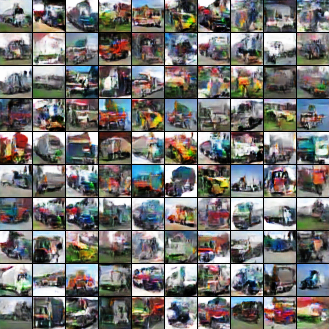}}
\caption{Samples of Triple-GAN-V1 in specific classes on the CIFAR10 dataset. The semantics of the samples are clear and correct in most of the cases.}
\label{fig:cifar_cc}
\end{figure}

\textbf{Class-conditional Generation.} We illustrate images generated by Triple-GAN-V1 in specific classes on CIFAR10 in Fig.~\ref{fig:cifar_cc}. In most cases, Triple-GAN-V1 is able to generate meaningful images with correct semantics.

\begin{figure}[t]
\centering
\subfigure[MNIST]{\includegraphics[height=.45\columnwidth]{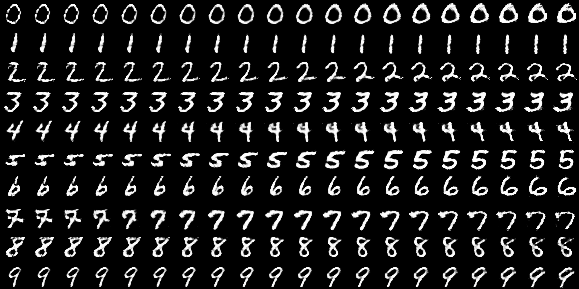}}
\subfigure[SVHN]{\includegraphics[height=.45\columnwidth]{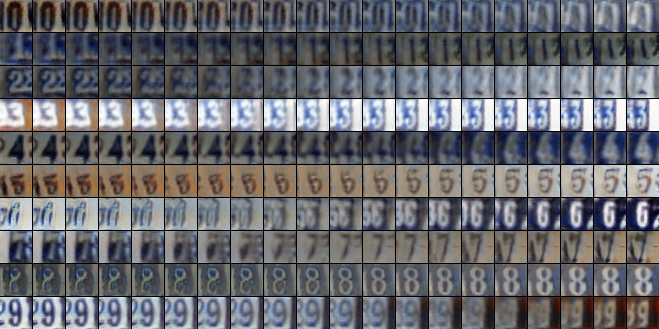}}
\subfigure[CIFAR10]{\includegraphics[height=.45\columnwidth]{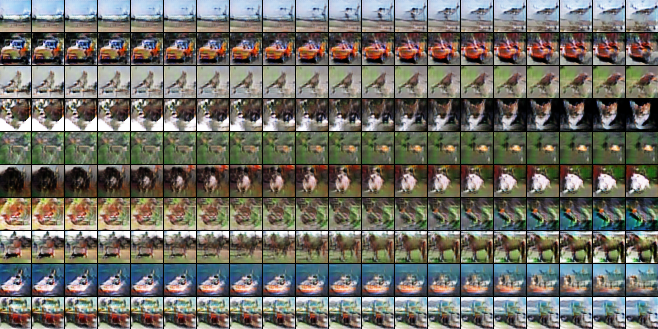}}
\caption{Class-conditional latent space interpolation. We first sample two random vectors in the latent space and interpolate linearly from one to another. Then, we map these vectors to the data level given a fixed label for each class. Totally, 20 images are shown for each class. We select two endpoints with clear semantics on the CIFAR10 dataset for better illustration.}
\label{fig:linear}
\end{figure}

\textbf{Latent Space Interpolation.} We demonstrate the generalization capability of our Triple-GAN-V1 via class-conditional latent space interpolation. As presented in Fig.~\ref{fig:linear}, Triple-GAN-V1 can transit smoothly from one sample to another with totally different visual factors without losing label semantics, which proves that Triple-GANs can learn meaningful latent spaces class-conditionally instead of overfitting to the training data, especially labeled data.

\section{Training Details}

We present the training details in both semi-supervised learning and the extreme low data regime.

\subsection{Semi-supervised learning}

In semi-supervised learning, we augment the labeled data in the training of $D$. In particular, we generate pseudo labels through $C$ for some unlabeled data and use these pairs as positive samples for $D$ to accept. Via this trick, $D$ will access samples beyond the empirical distribution of the small-size labeled data and encourage $G$ to generate diverse samples.
The cost of the trick is on introducing some bias to the target distribution of $D$, which is a mixture of $p_c(x, y)$ and $p(x, y)$ instead of the pure $p(x, y)$. However, this is acceptable as $C$ converges quickly and $p_c(x, y)$ and $p(x, y)$ are close.

\textbf{Triple-GAN-V1.} The pseudo discriminative loss is not applied until the number of epochs reaches a threshold that the generator could generate meaningful data. We search the threshold in $\{200, 300\}$, $\alpha_\mP$ in $\{0.1, 0.03\}$ and the global learning rate in $\{0.0003, 0.001\}$ based on the validation performance on each dataset. 
All of the other hyperparameters including relative weights and parameters in Adam~\cite{kingma2014adam} are fixed according to~\cite{salimans2016improved} across all of the experiments.
We list the detailed architectures of Triple-GAN-V1 on the MNIST, SVHN and CIFAR10 datasets in Tab.~\ref{table:a_m}, Tab.~\ref{table:a_s} and Tab.~\ref{table:a_c}, respectively. On Tiny ImageNet, we use the same architectures as on CIFAR10.

\begin{table*}[htbp]
	\centering
	\caption{ \textbf{Triple-GAN-V1 on MNIST}}
	\vspace{0.2cm}
	\begin{scriptsize}
	\begin{tabular}{c|c|c}
		\textbf{Classifier} C & \textbf{Discriminator} D & \textbf{Generator} G \bigstrut[b]\\
		\hline
		Input 28$\times$28 Gray Image & Input 28$\times$28 Gray Image, Ont-hot Class representation & Input Class y, Noise z \bigstrut\\
		\hline
		5$\times$5 conv. 32 ReLU &  MLP 1000 units, lReLU, gaussian noise, weight norm & MLP 500 units,  \bigstrut[t]\\
		2$\times$2 max-pooling, 0.5 dropout &  MLP 500 units, lReLU, gaussian noise, weight norm & softplus, batch norm \\
		3$\times$3 conv. 64 ReLU &  MLP 250 units, lReLU, gaussian noise, weight norm &  \\
		3$\times$3 conv. 64 ReLU &  MLP 250 units, lReLU, gaussian noise, weight norm & MLP 500 units, \\
		2$\times$2 max-pooling, 0.5 dropout &  MLP 250 units, lReLU, gaussian noise, weight norm &  softplus, batch norm \\
		3$\times$3 conv. 128 ReLU &  MLP 1 unit, sigmoid, gaussian noise, weight norm &  \\
		3$\times$3 conv. 128 ReLU &       & MLP 784 units, sigmoid \\
		Global pool &       &  \\
		10-class Softmax &       &  \\
	\end{tabular}%
	\end{scriptsize}
	\label{table:a_m}%
\end{table*}%

\begin{table*}[htbp]
	\centering
	\caption{\textbf{Triple-GAN-V1 on SVHN}}
	\vspace{0.2cm}
	\begin{scriptsize}
	\begin{tabular}{c|c|c}
		\textbf{Classifier} C & \textbf{Discriminator} D & \textbf{Generator} G \bigstrut[b]\\
		\hline
		Input: 32$\times$32 Colored Image & Input: 32$\times$32 colored image, class y & Input: Class y, Noise z \bigstrut\\
		\hline
		0.2 dropout & 0.2 dropout & MP 8192 units,  \bigstrut[t]\\
		3$\times$3 conv. 128 lReLU, batch norm & 3$\times$3 conv. 32, lReLU, weight norm & ReLU, batch norm \\
		3$\times$3 conv. 128 lReLU, batch norm & 3$\times$3 conv. 32, lReLU, weight norm, stride 2 & Reshape 512$\times$4$\times$4 \\
		3$\times$3 conv. 128 lReLU, batch norm &       & 5$\times$5 deconv. 256. stride 2, \\
		2$\times$2 max-pooling, 0.5 dropout & 0.2 dropout & ReLU, batch norm \bigstrut[b]\\
		\hline
		3$\times$3 conv. 256 lReLU, batch norm & 3$\times$3 conv. 64, lReLU, weight norm &  \bigstrut[t]\\
		3$\times$3 conv. 256 lReLU, batch norm & 3$\times$3 conv. 64, lReLU, weight norm, stride 2 &  \\
		3$\times$3 conv. 256 lReLU, batch norm &       & 5$\times$5 deconv. 128. stride 2,  \\
		2$\times$2 max-pooling, 0.5 dropout & 0.2 dropout & ReLU, batch norm \bigstrut[b]\\
		\hline
		3$\times$3 conv. 512 lReLU, batch norm & 3$\times$3 conv. 128, lReLU, weight norm &  \bigstrut[t]\\
		NIN, 256 lReLU, batch norm & 3$\times$3 conv. 128, lReLU, weight norm &  \\
		NIN, 128 lReLU, batch norm &       &  \\
		Global pool & Global pool & 5$\times$5 deconv. 3. stride 2,  \\
		10-class Softmax, batch norm & MLP 1 unit, sigmoid & sigmoid, weight norm \\
	\end{tabular}%
	\label{table:a_s}%
	\end{scriptsize}
\end{table*}%

\begin{table*}[htbp]
	\centering
	\caption{\textbf{Triple-GAN-V1 on CIFAR10}}
	\begin{scriptsize}
	\vspace{0.2cm}
	\begin{tabular}{c|c|c}
		\textbf{Classifier} C & \textbf{Discriminator} D & \textbf{Generator} G \bigstrut[b]\\
		\hline
		Input: 32$\times$32 Colored Image & Input: 32$\times$32 colored image, class y & Input: Class y, Noise z \bigstrut\\
		\hline
		Gaussian noise & 0.2 dropout & MLP 8192 units, \bigstrut[t]\\
		3$\times$3 conv. 128 lReLU, weight norm & 3$\times$3 conv. 32, lReLU, weight norm & ReLU, batch norm \\
		3$\times$3 conv. 128 lReLU, weight norm & 3$\times$3 conv. 32, lReLU, weight norm, stride 2 & Reshape 512$\times$4$\times$4 \\
		3$\times$3 conv. 128 lReLU, weight norm &       & 5$\times$5 deconv. 256.stride 2 \\
		2$\times$2 max-pooling, 0.5 dropout & 0.2 dropout & ReLU, batch norm \bigstrut[b]\\
		\hline
		3$\times$3 conv. 256 lReLU, weight norm & 3$\times$3 conv. 64, lReLU, weight norm &  \bigstrut[t]\\
		3$\times$3 conv. 256 lReLU, weight norm & 3$\times$3 conv. 64, lReLU, weight norm, stride 2 &  \\
		3$\times$3 conv. 256 lReLU, weight norm &       & 5$\times$5 deconv. 128. stride 2 \\
		2$\times$2 max-pooling, 0.5 dropout & 0.2 dropout & ReLU, batch norm \bigstrut[b]\\
		\hline
		3$\times$3 conv. 512 lReLU, weight norm & 3$,\times$3 conv. 128 lReLU, weight norm &  \bigstrut[t]\\
		NIN, 256 lReLU, weight norm & 3$\times$3 conv. 128, lReLU, weight norm &  \\
		NIN, 128 lReLU, weight norm &       &  \\
		Global pool & Global pool & 5$\times$5 deconv. 3. stride 2 \\
		10-class Softmax with weight norm & MLP 1 unit, sigmoid, weight norm & tanh, weight norm \\
	\end{tabular}%
	\label{table:a_c}%
	\end{scriptsize}
\end{table*}%

\begin{table*}[htbp]
	\centering
	\caption{\textbf{Triple-GAN-V2 on SVHN, CIFAR10 and Tiny ImageNet}}
	\begin{scriptsize}
	\vspace{0.2cm}
	\begin{tabular}{c|c|c}
		\textbf{Classifier} C & \textbf{Discriminator} D & \textbf{Generator} G \bigstrut[b]\\
		\hline
		Input: 32$\times$32 Colored Image & Input: 32$\times$32 colored image, class y & Input: Class y, Noise z \bigstrut\\
		Randomly $\{\Delta x, \Delta y \}$  $\in [-2, 2]$  &\\
		Randomly Flip with $p=0.5$ (except on SVHN)  &\\ 
		\hline
		Gaussian noise & 3$\times$3 2-layer residual block. 2048. stride 2 & MLP 8$\times$256$\times$4 units, \bigstrut[t]\\
		3$\times$3 conv. 128 lReLU, weight norm & ReLU, spectral norm &  Reshape 2048$\times$2$\times$2\\
		3$\times$3 conv. 128 lReLU, weight norm &  3$\times$3 2-layer residual block. 2048. stride 2  & 3$\times$3 2-layer residual block. 2048. stride 2\\
		3$\times$3 conv. 128 lReLU, weight norm &       & \\
		2$\times$2 max-pooling, 0.5 dropout & ReLU, spectral norm & ReLU, batch norm \bigstrut[b]\\
		\hline
		3$\times$3 conv. 256 lReLU, weight norm &  3$\times$3 2-layer residual block. 2048. stride 2  &  \bigstrut[t]  3$\times$3 2-layer residual block. 1024.stride 2 \\
		3$\times$3 conv. 256 lReLU, weight norm & ReLU, spectral norm &  ReLU, batch norm\\
		3$\times$3 conv. 256 lReLU, weight norm &      3$\times$3 2-layer residual block. 2048. stride 2   &  3$\times$3 2-layer residual block. 512. stride 2\\
		2$\times$2 max-pooling, 0.5 dropout & ReLU, spectral norm & ReLU, batch norm \bigstrut[b]\\
		\hline
		3$\times$3 conv. 512 lReLU, weight norm & MLP 2048 units &  \bigstrut[t]\\
		NIN, 256 lReLU, weight norm & ReLU, spectral norm &  \\
		NIN, 128 lReLU, weight norm &   MLP 10 units    &  \\
		Global pool & & 3$\times$3 2-layer residual block. 256. stride 2 \\
		10-class Softmax with weight norm & 10-class Softmax, spectral norm  & tanh \\
	\end{tabular}%
	\label{table:tgan_v2}%
	\end{scriptsize}
\end{table*}%

\begin{table*}[htbp]
	\centering
	\caption{\textbf{Triple-GAN-V2 with a 26-layer ResNet classifier on CIFAR10}}
	\begin{scriptsize}
	\vspace{0.2cm}
	\begin{tabular}{c|c|c}
		\textbf{Classifier} C & \textbf{Discriminator} D & \textbf{Generator} G \bigstrut[b]\\
		\hline
		Input: 32$\times$32 Colored Image & Input: 32$\times$32 colored image, class y & Input: Class y, Noise z \bigstrut\\
		Randomly $\{\Delta x, \Delta y \}$  $\in [-4, 4]$ &\\
		Randomly Flip with $p=0.5$  &\\ 
		\hline
		Gaussian noise & 3$\times$3 2-layer residual block. 2048. stride 2 & MLP 8$\times$256$\times$4 units, \bigstrut[t]\\
		3$\times$3 conv. 16. & ReLU, spectral norm &  Reshape 2048$\times$2$\times$2\\
		ReLU, batch norm &  3$\times$3 2-layer residual block. 2048. stride 2  & 3$\times$3 2-layer residual block. 2048. stride 2\\
	    3$\times$3 conv. 4-layer ShakeShake block. 96.  &       & \\
	    ReLU, batch norm  & ReLU, spectral norm & ReLU, batch norm \bigstrut[b]\\
		\hline
		 3$\times$3 conv. 4-layer ShakeShake block. 192. stride 2 &  3$\times$3 2-layer residual block. 2048. stride 2  &  \bigstrut[t]  3$\times$3 2-layer residual block. 1024.stride 2 \\
		ReLU, batch norm  & ReLU, spectral norm &  ReLU, batch norm\\
		3$\times$3 conv. 4-layer ShakeShake block. 384. stride 2 &      3$\times$3 2-layer residual block. 2048. stride 2   &  3$\times$3 2-layer residual block. 512. stride 2\\
		ReLU, batch norm & ReLU, spectral norm & ReLU, batch norm \bigstrut[b]\\
		\hline
		Global pool & MLP 2048 units &  \bigstrut[t]\\
		3$\times$3 conv. 4-layer ShakeShake block. 384. & ReLU, spectral norm &  \\
		MLP 10 units &   MLP 10 units    &  \\
		 &  & 3$\times$3 2-layer residual block. 256. stride 2 \\
		10-class Softmax & 10-class Softmax, spectral norm & tanh\\
	\end{tabular}%
	\label{table:tgan_v2_resnet}%
	\end{scriptsize}
\end{table*}%

\textbf{Triple-GAN-V2.} We use the exact same classifier architecture, data augmentation protocol and hyperparameters as in the baseline~\cite{tarvainen2017mean} unless specified. In particular, the coefficient $\alpha_{\mU}$ and learning rate have a ramp-up period of 40,000 iterations from the beginning and they ramped up from 0 to the maximum values (50.0 for $\alpha_{\mU}$ and 0.001 for the learning rate), using a
sigmoid-shaped function $e^{-5(1-x)^2}$, where $x \in [0, 1]$. On Tiny ImageNet, 
Triple-GAN-V2 is trained by 100,000 iterations in total and the first 10,000 iterations are for pretraining of the classifier. On other datasets, Triple-GAN-V2 is trained by 220,000 iterations in total and the first 20,000 iterations are for pretraining of the classifier.
The pseudo discriminative loss is not applied until $50,000$ iterations on SVHN and CIFAR10 and $30,000$ iterations on Tiny ImageNet. We also add a ramp-up period of 20,000 iterations steps for $\alpha_\mP$ and it ramped up from 0 to $3.0$, using the same ramp-up function as $\alpha_{\mU}$. We use a projection discriminator~\cite{miyato2018cgans} with spectral normalization~\cite{miyato2018spectral}.
All of the other hyperparameters including relative weights and parameters in Adam~\cite{kingma2014adam} are fixed according to~\cite{tarvainen2017mean,miyato2018cgans} across all of the experiments.
We list the detailed architectures of Triple-GAN-V2 and Triple-GAN-V2 with a 26-layer ResNet classifier in Tab.~\ref{table:tgan_v2} and Tab.~\ref{table:tgan_v2_resnet}, respectively.

\subsection{Extreme low data regime}

All architectures are the same in both semi-supervised learning and the extreme low data regime on the same dataset. Triple-GAN-V2 is trained by 100,000/150,000/100,000 iterations in total with the maximum value of $\alpha_\mP$ to be 0.01/0.3/0.01 on SVHN, CIFAR10 and Tiny ImageNet. Other hyperparameters are kept the same as in the semi-supervised learning. We also add the balanced consistency regularization on GANs~\cite{zhao2020improved} to relieve the mode collapse and the related hyperparameters are set exactly the same as the original paper. See more details in our source code.

\end{document}